\documentclass{article}
\usepackage{titlesec}
\usepackage{cite}
\usepackage{amsmath, amssymb}
\usepackage{enumitem}
\usepackage{authblk}
\usepackage{geometry}
\usepackage{subfigure,graphicx}
\usepackage[hyperfootnotes=false]{hyperref}
\usepackage{multirow}
\usepackage{booktabs}
\usepackage{amsfonts}
\usepackage{algorithm}
\usepackage{stmaryrd}
\usepackage{algpseudocode}
\usepackage{array}
\usepackage{float}
\usepackage{stmaryrd}
\usepackage{textcomp}
\usepackage{ccicons}
\title{\textbf{Opus} \vspace{0.5em} \\ \Large{\fontseries{bx} A Workflow Intention Framework for Complex Workflow Generation}}

\author{
\vspace{1em}
\begin{minipage}[t]{0.33\textwidth}
    \centering
    \begin{tabular}[t]{c}
        \small{\text{\textbf{Phillip Kingston}$^1$}}\\ \vspace{-0.7em}
        \scriptsize{Member of Technical Staff} \\
        \scriptsize{AppliedAI} \\
    \end{tabular}
\end{minipage}%
\begin{minipage}[t]{0.33\textwidth}
    \centering
    \begin{tabular}[t]{c}
        \small{\text{\textbf{Théo Fagnoni}$^1$}} \\ \vspace{-0.7em}
        \scriptsize{Member of Technical Staff} \\
        \scriptsize{AppliedAI} \\
    \end{tabular}
\end{minipage}%
\begin{minipage}[t]{0.33\textwidth}
    \centering
    \begin{tabular}[t]{c}
        \small{\text{\textbf{Mahsun Altin}$^2$}} \\ \vspace{-0.7em}
        \scriptsize{Member of Technical Staff} \\
        \scriptsize{AppliedAI} \\
    \end{tabular}
\end{minipage}%
\vspace{1em}
}

\date{\normalsize{25 January 2025}}

\begin{document}

\maketitle

\begin{center}
    \ccbyncsa \\
    \vspace{0.3em}
    \footnotesize{This work is licensed under a Creative Commons Attribution-Noncommercial-ShareAlike 4.0 International License (CC BY-NC-SA 4.0)}
\end{center}

\footnotetext{\text{$^1$ Equal contribution, corresponding author: phillip.kingston@opus.com}}
\footnotetext{\text{$^2$ Minor contribution}}

\vspace{1em}

\begin{abstract}

\noindent This paper introduces Workflow Intention, a novel framework for identifying and encoding process objectives within complex business environments. Workflow Intention is the alignment of Input, Process and Output elements defining a Workflow’s transformation objective interpreted from Workflow Signal inside Business Artefacts. It specifies how Input is processed to achieve desired Output, incorporating quality standards, business rules, compliance requirements and constraints. We adopt an end-to-end Business Artefact Encoder and Workflow Signal interpretation methodology involving four steps: Modality-Specific Encoding, Intra-Modality Attention, Inter-Modality Fusion Attention then Intention Decoding. We provide training procedures and critical loss function definitions. In this paper:

\begin{enumerate}
    \item We introduce the concepts of Workflow Signal and Workflow Intention, where Workflow Signal—decomposed into Input, Process and Output elements—is interpreted from Business Artefacts, and Workflow Intention is a complete triple of these elements.
    \item We introduce a mathematical framework for representing Workflow Signal as a vector and Workflow Intention as a tensor, formalizing properties of these objects.
    \item We propose a modular, scalable, trainable, attention-based multimodal generative system to resolve Workflow Intention from Business Artefacts.
\end{enumerate}

\end{abstract}

\newpage

\section{Introduction}
In the lifecycle of most businesses and government organizations, developing and refining internal processes is crucial for maintaining structure, consistency, and overall operational efficiency. Well-defined processes offer tangible advantages: they enhance quality control, reduce costs, mitigate risks, and facilitate auditing. They also help ensure continuity when employees retire or transition out of the organization. Furthermore, many regulatory and compliance agencies publish procedural guidelines to help external stakeholders understand and meet specific standards.\\

Despite the clear benefits of process documentation, its quality, completeness, currency, accuracy, and granularity can vary widely across organizations. This is where the concept of Workflow Intention becomes invaluable. By extracting the core purpose and objectives from existing Business Artefacts—such as standard operating procedures, policy manuals, and historical records—Workflow Intention enables organizations to rapidly implement supervised automation and evolve legacy processes into efficient AI-enhanced Workflows enriched with best practice.

\paragraph{Definitions}

Our methodology is based on the following concepts introduced in \textit{Opus: A Large Workflow Model for Complex Workflow Generation} by Fagnoni et al. \cite{opuslargeworkmodel}:

\vspace{1em}

\noindent\hangindent=2em\hangafter=0 \textbf{\textbf{Input}:} The dataset initiating a \textbf{Process}, conforming to validation rules and format specifications. \textbf{Input} is multimodal, including structured (e.g. databases, forms) and unstructured (e.g. documents, media) data types such as text, documents, images, audio, and video.

\vspace{1em}

\noindent\hangindent=2em\hangafter=0 \textbf{Process:} A structured sequence of operational steps transforming \textbf{Input} into \textbf{Output}, defined in part or whole across \textbf{Business Artefacts}. \textbf{Process} combines automated and manual steps defining start/end conditions, decision points, parallel paths, roles, success criteria, monitoring, metrics, compliance requirements and error handling.

\vspace{1em}

\noindent\hangindent=2em\hangafter=0 \textbf{Output:} The result of a \textbf{Process} operating on \textbf{Input}, meeting predefined quality and business criteria. \textbf{Output} can be tangible (e.g. documents) or intangible (e.g. decisions) and include audit trails of their creation. Supported formats include text and documents.

\vspace{1em}

\noindent\hangindent=2em\hangafter=0 \textbf{Business Artefact:} Any text, document, image, audio, or video capturing \textbf{Process} knowledge (e.g. process maps, standard operating procedures, regulatory guidelines, compliance documents) or that provides  \textbf{Workflow Signals}. \textbf{Business Artefacts} detail input/output specifications, task sequences, business rules and roles.

\vspace{1em}

\noindent\hangindent=2em\hangafter=0 \textbf{Workflow:} A software defined executable \textbf{Process} as a sequence of \textbf{Tasks}. \textbf{Workflows} coordinate task execution, manage data flow, and enforce business rules, compliance, and process logic (e.g. conditionals, loops, error handling). \textbf{Workflows} support monitoring, logging, audit, state management, concurrency, adaptive modification, and version control.

\vspace{1em}

\noindent\hangindent=2em\hangafter=0 \textbf{Task:} An atomic unit of work within a \textbf{Workflow}, performing a specific function with defined input/output schemas, objectives, timing constraints, and success criteria. \textbf{Tasks} adhere to the singular responsibility principle, support automation or manual intervention, and maintain contextual awareness of dependencies. \textbf{Tasks} are auditable by humans or AI agents against their definition.

\vspace{1em}

\noindent\hangindent=2em\hangafter=0 \textbf{Workflow Intention:} The alignment of \textbf{Input}, \textbf{Output}, and \textbf{Process} components defining a \textbf{Workflow}'s transformation objective. It specifies how \textbf{Input} is processed to achieve desired \textbf{Output}, incorporating data formats, quality standards, business rules, and constraints. It is determined by interpreting \textbf{Workflow Signals} from direct and indirect sources.

\vspace{1em}

\noindent\hangindent=2em\hangafter=0 \textbf{Workflow Signal:} A discrete informational cue from \textbf{Business Artefacts} or \textbf{Intention Elicitation} that conveys implicit or explicit information on \textbf{Input}, \textbf{Process} or \textbf{Output} relevant to a \textbf{Workflow}.

\vspace{1em}

\noindent\hangindent=2em\hangafter=0 \textbf{Intention Elicitation:} User-driven communication (e.g. text-based conversations, interface interactions) that contains \textbf{Workflow Signals} to further articulate \textbf{Workflow Intention(s)}. It captures objectives, constraints, and preferences, distinct from \textbf{Business Artefacts} and \textbf{Input}/\textbf{Output} examples.

\vspace{1em}

\noindent\hangindent=2em\hangafter=0 \textbf{Complete Intention:} The state where sufficient information exists across \textbf{Input}, \textbf{Output}, and \textbf{Process} components for accurate \textbf{Workflow} implementation. Incomplete intentions lack clear specifications, relationships, or operational requirements, hindering execution.

\vspace{1em}

\noindent\hangindent=2em\hangafter=0 \textbf{Mixed Intention:} The state where \textbf{Business Artefacts} or \textbf{Intention Elicitation} describe multiple distinct transformation objectives, requiring separation into individual \textbf{Workflow Intentions}. Separation improves clarity, maintainability, and preserves \textbf{Workflow} interfaces.

\vspace{1em}

This paper makes the following key contributions:
\begin{enumerate}
    \item We introduce the concepts of Workflow Signal and Workflow Intention. Workflow Signals are interpreted from Business Artefacts and decomposed into Input ($i$), Process ($p$) and Output ($o$) elements. Workflow Intention is defined as a triple of Workflow Signals $i$, $p$ and $o$. 
    \item We introduce a mathematical framework for representing Workflow Signals as vectors $i, p, o$ and Workflow Intention as a tensor $(i, p, o)$, formalizing properties of these objects under this framework.
    \item We propose a modular, scalable, trainable, attention-based multimodal generative system to resolve Workflow Intention.
\end{enumerate}

\section{Background}
This framework leverages many state-of-the-art methodologies to obtain contextual representations of multimodal Business Artefacts in order to generate Workflow Intention. Transformer-based architectures  introduced a paradigm shift from recurrence to parallelizable self-attention, enabling efficient long-range dependency modeling and significantly improving scalability in natural language processing. Subsequent innovations in encoder-only (RoBERTa), decoder-only (GPT), and encoder-decoder (T5) architectures further refined contextual representation and generation capabilities. Advancements in visual (ViT) and multimodal (FLAVA, NVLM, InternVL) transformers have broadened these capabilities to handle image and text jointly, providing robust backbones for extracting structured features from Business Artefacts to facilitate Workflow Intention generation.

\paragraph{Transformer Architecture}
The transformer model introduced by Vaswani et al. \cite{attention} replaced recurrence with self-attention, allowing tokens to attend globally within a sequence. This enabled parallel processing and improved long-range dependency modeling, with the multi-head attention mechanism capturing diverse relationships. By eliminating step-by-step processing, transformers became highly scalable and efficient for NLP tasks. This mechanism is central to our approach as we aim to extract dense features from long-range sequences produced by Business Artefacts.

\paragraph{Encoder, Decoder, Encoder-Decoder}
RoBERTa by Liu et al. \cite{roberta} is an optimized encoder-only architecture designed to enhance contextual representations in language models. It refines pretraining strategies by removing next sentence prediction, incorporating dynamic masking and extending training on larger datasets. As an encoder-based model, it effectively captures rich representations, which we leverage to preserve features consistently across our Workflow Intention generation pipeline.\\ 

Decoder-only architectures, such as GPT, rely exclusively on a causal decoder to generate text recursively. Unlike encoder models, GPT processes input unidirectionally, meaning tokens attend only to previous tokens, ensuring autoregressive generation. The decoder generates text one token at a time, using a stopping mechanism to determine completion. We employ a decoder to generate vectors based on our framework to represent Workflow Intention.\\  

T5 by Raffel et al. \cite{t5} extends transformer capabilities by combining an encoder and decoder with cross-attention, where the decoder attends to encoded representations before generating output. Unlike GPT, which generates step by step based on past tokens, T5 benefits from a bidirectional encoder, capturing full context before passing information to the decoder. This architecture allows us to generate Workflow Intention based on a context captured from Business Artefacts.

\paragraph{Visual Transformers}
The vision transformer adapted transformers for images by dividing input images into fixed-size patches, treating them as tokens, and applying self-attention. This allowed the model to capture both local and global relationships without convolutions. ViT by Dosovitskiy et al. \cite{vit} demonstrated that a pure attention-based approach can match or surpass CNNs on vision tasks when trained on large datasets, proving the generalizability of transformers beyond NLP. We leverage these architectures to process document and image Business Artefacts.

\paragraph{Multimodal Transformers}
Multimodal transformers integrate text and vision, allowing AI models to reason across modalities. FLAVA by Singh et al. \cite{flava} combines separate image and text encoders with a multimodal encoder to align representations for captioning and visual question-answering tasks. T5-inspired architectures for vision-language tasks leverage cross-attention to fuse textual and visual embeddings effectively. NVLM by Dai et al. \cite{nvidiaNVLM}, a large multimodal LLM, integrates vision encoders into an LLM while maintaining strong language capabilities, excelling at tasks requiring both modalities. InternVL by Chen et al. \cite{internVL} scales multimodal learning further, progressively aligning large vision models with text models to handle diverse inputs, including video and complex multimodal reasoning. These state-of-the-art scalable architectures serve as backbones for ingesting multimodal Business Artefacts, constructing context from them, and generating Workflow Intention.

\section{System Overview}

Let \( \mathcal{M} = \{m_1, m_2, \dots, m_K\} \) denote the set of \( K \) distinct modalities. For each modality \( m_k \), we consider a set of Business Artefacts \( \mathcal{A}_{m_k} = \{a_{m_k,1}, a_{m_k,2}, \dots, a_{m_k,N_{m_k}}\} \), where \( N_{m_k} \) is the number of Business Artefacts in modality \( m_k \). Each Business Artefact \( a_{m_k,i} \) is represented in its raw form, with modality-specific dimensionality. In this paper, we consider three modalities: Text ($\text{T}$), Image ($\text{I}$) and Document ($\text{D}$). The framework is built to support any modality. \\

Each Business Artefact from $\mathcal{A}_{m_k}$ gets encoded in a dedicated pipeline. Then, all encoded Business Artefacts from $\mathcal{A}_{m_k}$ are concatenated and encoded by a dedicated intra-modality pipeline.
The encoded Business Artefacts are then concatenated across the modalities by a fusion encoder, before entering the Workflow Intention decoder. The encoded decoder generates vectors which are projected into Workflow Intention objects until it stops.\\

\begin{figure}[H]
    \centering
    \includegraphics[width=\textwidth]{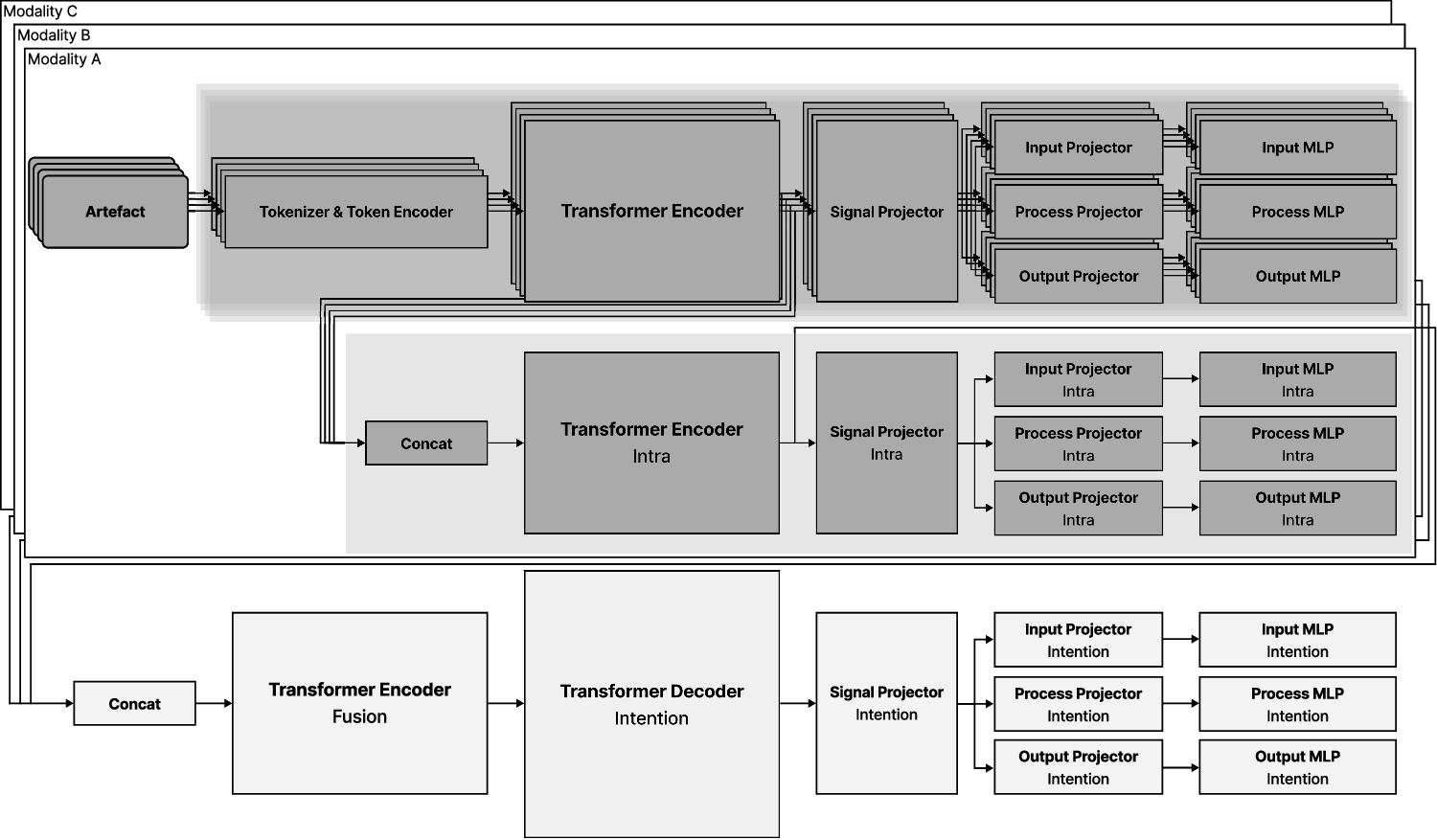}
    \caption{High-Level System Overview}
    \label{fig:system}
\end{figure}

\newpage

\section{Business Artefacts Encoding and Signal Extraction}

Each Business Artefact $a_{m_k, i}$ of modality $m_k$ is tokenized by a tokenizer $\textbf{Tok}_{m_k}$. Each token is embedded by a token encoder $\mathbf{E}_{m_k}$. The resulting sequence of vectors $\text{E}_{m_k, i}$ is encoded by a self-attention based encoder $\textbf{Encoder}_{m_k}$ into $\text{E}^{\text{enc}}_{m_k, i}$. 
The encoded sequence is  projected into a unified space via a linear projection parametrized by $\text{W}^u_{m_k}$, $b^u_{m_k}$ and the arrival dimension $d$, giving $\text{H}_{m_k, i}$. 
A learned representative vector $h^{\texttt{[REP]}}_{i, m_k}$ is retrieved or computed from $\text{H}_{m_k, i}$ and linearly projected by three Projection Heads: Input ($\text{W}^{\text{I}}_{m_k}$, $b^{\text{I}}_{m_k}$), Process ($\text{W}^{\text{P}}_{m_k}$, $b^{\text{P}}_{m_k}$) and Output ($\text{W}^{\text{O}}_{m_k}$, $b^{\text{O}}_{m_k}$).

\subsection{Modality-Specific Business Artefact Encoding}
\subsubsection{Text Business Artefacts}
Let T be the text modality. For ease of notation, let $T_i = a_{\text{T},i}$ a text Business Artefact.
\paragraph{Text Tokenizer and Token Encoder}
We tokenize $T_i$ with {$\textbf{Tok}_\text{T}$ into a sequence of \( L_{T_i}\) tokens $T_i \mapsto \{\texttt{[CLS]}_\text{T}, t_{i,1}, t_{i,2}, \dots, t_{i,L_{T_i} - 1} \}$, prepending a $\texttt{[CLS]}_\text{T}$ token. Each token \( t_{i,j} \) is mapped to an embedding vector $\mathbf{E}_\text{T}(t_{i,j}) \in \mathbb{R}^{d_\text{T}} $ by a text token encoder $\mathbf{E}_\text{T}: \mathcal{V}_\text{T} \to \mathbb{R}^{d_\text{T}}$} where \(\mathcal{V}_\text{T}\) is the vocabulary (the set of all possible text tokens) and $d_\text{T}$ is the embedding dimensionality. We define the tensor representation $\text{E}_{T_i} = \text{E}_{\text{T}, i}$ for ease of notation as the sequence of embedded tokens:
\begin{equation}
\text{E}_{T_i} = \big[ \mathbf{E_\text{T}}(\texttt{[CLS]}_\text{T}), \mathbf{E_\text{T}}(t_{i,1}),\, \mathbf{E_\text{T}}(t_{i,2}),\, \dots,\, \mathbf{E_\text{T}}(t_{i,L_{T_i}-1}) \big] \in \mathbb{R}^{d_\text{T} \times L_{T_i}}
\end{equation}

\paragraph{Text Encoder}

To capture contextual dependencies across all tokens in \(T_i\), we pass \(\text{E}_{T_i}\) through an encoder network $\textbf{Encoder}_{\text{T}}$ to calculate $\text{E}^{\text{enc}}_{T_i} = \text{E}^{\text{enc}}_{\text{T},i}$ in which each column contains a contextualized embedding of the corresponding token. Finally, we employ a linear projection to map $\text{E}^{\text{enc}}_{T_i}$ into a \(d\)-dimensional space. The projection is achieved by applying a learnable weight matrix \(\text{W}^u_\text{T} \in \mathbb{R}^{d \times d_T}\) and a bias vector \(b^u_\text{T} \in \mathbb{R}^{d}\), thus forming the final representation \(\text{H}_{T_i}\):

\begin{equation}
\text{H}_{T_i} = \text{W}^u_\text{T} \text{E}^{\text{enc}}_{T_i} + b^u_\text{T}, \quad  \text{H}_{T_i} \in \mathbb{R}^{d\times L_{T_i}}
\end{equation}

\subsubsection{Image Business Artefacts}
Let F be the image modality. For ease of notation, let $F_i = a_{\text{F},i}$ an image Business Artefact. $F_i$ is represented in its raw form as a three-dimensional tensor of size \(\mathbb{R}^{c \times h \times w}\), where \(c\) denotes the number of channels, \(h\) and \(w\) represent the height and width of the image, respectively.

\paragraph{Image Tokenizer and Token Encoder}
To remain consistent with the tokenizing abstraction, i.e. ``tokenizer” for text, we adopt the same terminology, even though it is actually a specific image feature extraction described in the appendix.\\

We tokenize $F_i$ with $\textbf{Tok}_\text{F}$ into a sequence of \( L_{F_i}\) patches $F_i \mapsto \{f_{i,1}, f_{i,2}, \dots, f_{i,L_{F_i}} \}$. We do not employ any $\texttt{[CLS]}$-type vector representation, consistent with ViT (Dosovitskiy et al. \cite{vit}). Each patch \( f_{i,j} \) is mapped to an embedding vector $\mathbf{E}_\text{F}(f_{i,j}) \in \mathbb{R}^{d_\text{F}} $ by an image patch encoder $\mathbf{E}_\text{F}: \mathbb{R}^{c \times h \times w} \to \mathbb{R}^{d_\text{F}}$ where $d_\text{F}$ is the embedding dimensionality.\\

We define the tensor representation $\text{E}_{F_i} = \text{E}_{\text{F}, i}$ for ease of notation as the sequence of embedded patches:
\begin{equation}
\text{E}_{F_i} = \big[ \mathbf{E_\text{F}}(f_{i,1}),\, \mathbf{E_\text{F}}(f_{i,2}),\, \dots,\, \mathbf{E_\text{F}}(f_{i,L_{F_i}}) \big] \in \mathbb{R}^{d_\text{F} \times L_{F_i}}
\end{equation}

\paragraph{Image Encoder}
To capture contextual dependencies across all patches in \(F_i\), we pass \(\text{E}_{F_i}\) through an encoder network $\textbf{Encoder}_{\text{F}}$ to calculate $\text{E}^{\text{enc}}_{F_i} = \text{E}^{\text{enc}}_{\text{F},i}$ in which each column contains a contextualized embedding of the corresponding token. Finally, we employ a linear projection to map $\text{E}^{\text{enc}}_{F_i}$ into a \(d\)-dimensional space. The projection is achieved by applying a learnable weight matrix \(\text{W}^u_\text{F} \in \mathbb{R}^{d \times d_F}\) and a bias vector \(b^u_\text{F} \in \mathbb{R}^{d}\), thus forming the final representation \(\text{H}_{F_i}\):

\begin{equation}
\text{H}_{F_i} = \text{W}^u_\text{F} \text{E}^{\text{enc}}_{F_i} + b^u_\text{F}, \quad  \text{H}_{F_i} \in \mathbb{R}^{d\times L_{F_i}}
\end{equation}

\subsubsection{Document Business Artefacts}
Let D be the document modality. We treat each document page as an image of size $(h_D, w_D)$. Each page is tokenized separately.
For ease of notation, let $D_{p,i} = a_{\text{D}_p,i}$ a page of a document Business Artefact and $D_i = a_{\text{D},i}$ a document Business Artefact.

\paragraph{Document Page Tokenizers and Token Encoders}
$D_{p,i}$ is decomposed as (inspired by the method of NVLM by Dai et al. \cite{nvidiaNVLM}):
\begin{itemize}
   \item Text elements $\{D^{\text{T}}_{p, i, q}\}_q$: a set of text elements, each tokenized through $\textbf{Tok}_{\text{T}}$ then concatenated, prepended with a $\texttt{[CLS]}_\text{T}$ token, producing a sequence of $L_{D_{p,i}}^{\text{T}}$ tokens $\{d^\text{T}_{i,j}\}_j$. Each token \( d^\text{T}_{i,j} \) is mapped to an embedding vector \(\mathbf{E}_\text{T}(d^\text{T}_{i,j}) \in \mathbb{R}^{d_\text{T}} \) using $\mathbf{E}_\text{T}: \mathcal{V}_\text{T} \to \mathbb{R}^{d_\text{T}}$, producing a sequence of vectors denoted by $\text{E}^{\text{T}}_{D_{p,i}} \in \mathbb{R}^{d_\text{T} \times L_{D_{p,i}}^{\text{T}}}$.
   
   \item Text spatial elements: bounding box coordinates of each text token $d^\text{T}_{i,j}$ expressed in text as $\texttt{box}^\text{T}_{i,j}$ = ``${\texttt{<box>}(x^{\text{T}}_{\min, i,j}, y^{\text{T}}_{\min, i,j}), (x^{\text{T}}_{\max, i,j}, y^{\text{T}}_{\max, i,j}) \texttt{</box>}}$'', with \\$\texttt{box}^\text{T}_{i,\texttt{[CLS]}_\text{T}}$ = ``$\texttt{<box>}((0, 0), (h_D, w_D))\texttt{</box>}$'', and mapped to an embedding \( \mathbf{E}_\text{S}(\texttt{box}^\text{T}_{i,j}) \in \mathbb{R}^{d_{\text{T}}} \), producing a sequence of vectors denoted by $\text{E}^{\text{T}_s}_{D_{p,i}} \in \mathbb{R}^{d_\text{T} \times L_{D_{p,i}}^{\text{T}}}$. $\mathbf{E}_\text{S} = \overline{\textbf{Encoder}_{\text{T}}}$ denotes the average over the sequence of embeddings produced by the text encoder, in order to obtain one embedding per bounding box.

   \item Image elements $\{D^{\text{F}}_{i, q}\}_q$: a set of image elements, each patched through $\textbf{Tok}_{\text{F}}$ then concatenated, producing a sequence of $L_{D_{p,i}}^{\text{F}}$ patches $\{d^\text{F}_{i,j}\}_j$. Each patch \( d^\text{F}_{i,j} \) is mapped to an embedding vector \(\mathbf{E}_\text{F}(d^\text{F}_{i,j}) \in \mathbb{R}^{d_\text{F}} \) using $\mathbf{E}_\text{F}: \mathbb{R}^{c \times h \times w} \to \mathbb{R}^{d_\text{F}}$, producing a sequence of vectors denoted by $\text{E}^{\text{F}}_{D_{p,i}} \in \mathbb{R}^{d_\text{F} \times L_{D_{p,i}}^{\text{F}}}$.

   \item Image spatial elements: bounding box coordinates of each image patch $d^\text{F}_{i,j}$
   expressed in text as $\texttt{box}^\text{F}_{i,j}$ = ``${\texttt{<box>}(x^{\text{F}}_{\min, i,j}, y^{\text{F}}_{\min, i,j}), (x^{\text{F}}_{\max, i,j}, y^{\text{F}}_{\max, i,j}) \texttt{</box>}}$'', with \\ $\texttt{box}^\text{F}_{i,\texttt{[CLS]}_\text{F}}$ = ``$\texttt{<box>}(0, 0), (h_D, w_D))\texttt{</box>}$'', and mapped to an embedding \( \mathbf{E}_\text{S}(\texttt{box}^\text{F}_{i,j}) \in \mathbb{R}^{d_{\text{T}}} \), producing a sequence of vectors denoted by $\text{E}^{\text{F}_s}_{D_{p,i}} \in \mathbb{R}^{d_\text{T} \times L_{D_{p,i}}^{\text{F}}}$. $\mathbf{E}_\text{S} = \overline{\textbf{Encoder}_{\text{T}}}$ denotes the average over the sequence of embeddings produced by the text encoder, in order to obtain one embedding per bounding box.

\end{itemize}

Since the dimensions \( d_\text{T} \) and \( d_\text{F} \) may differ, we project each patch embedding into  $\mathbb{R}^{d_\text{T}}$ via a learnable linear projection with bias term $(\text{W}^\text{D}_\text{F} \in \mathbb{R}^{d_\text{T} \times d_\text{F}}$, $b^\text{D}_\text{F} \in \mathbb{R}^{d_\text{T}})$ to obtain 

\begin{equation}
\tilde{\text{E}}^{\text{F}}_{D_{p,i}} = \text{W}^\text{D}_\text{F}\text{E}^{\text{F}}_{D_{p,i}} + b^\text{D}_\text{F}\in \mathbb{R}^{d_\text{T} \times L_{D_{p,i}}^{\text{F}}}
\end{equation}

Document text and image element embeddings are concatenated with their respective spacial element embeddings to produce a sequence of vectors as follows:

\begin{equation}
\text{E}_{D_{p,i}} = \text{Concat}(\text{E}^\text{F, Concat}_{D_{p,i}}, \text{E}^\text{T, Concat}_{D_{p,i}}) \in \mathbb{R}^{d_\text{T} \times  L_{D_{p,i}}} \text{ with } L_{D_{p,i}} = 2(L_{D_{p,i}}^{\text{F}} + L_{D_{p,i}}^{\text{T}})
\end{equation}

where 
\begin{align}
    \text{E}^\text{F, Concat}_{D_{p,i}} &= \text{Concat}((\text{W}^\text{D}_\text{F}\mathbf{E}_\text{F}(d^\text{F}_{i,j}) + b^\text{D}_\text{F}, \mathbf{E}_\text{S}(\texttt{box}^\text{F}_{i,j}))_j) \in \mathbb{R}^{d_\text{T} \times  2L_{D_{p,i}}^{\text{F}}} \\
    \text{E}^\text{T, Concat}_{D_{p,i}} &= \text{Concat}((\mathbf{E}_\text{T}(d^\text{T}_{i,j}), \mathbf{E}_\text{S}(\texttt{box}^\text{T}_{i,j}))_j) \in \mathbb{R}^{d_\text{T} \times  2L_{D_{p,i}}^{\text{T}}}
\end{align}

\paragraph{Document Encoder}
The tokenized pages $\{{E}_{D_{p,i}, n}\}_n$ are concatenated into $\text{E}_{D_{i}}$.
To capture contextual dependencies across all tokens and patches in \(D_i\), we pass \(\text{E}_{D_{i}}\) through an encoder network 
$\textbf{Encoder}_{\text{D}}$ to calculate $\text{E}^{\text{enc}}_{D_{i}} = \text{E}^{\text{enc}}_{\text{D},i}$ in which each column contains a contextualized embedding of the corresponding token, token spatial coordinates, patch and patch spatial coordinates. Finally, we employ a linear projection to map $\text{E}^{\text{enc}}_{D_{i}}$ into a \(d\)-dimensional space. The projection is achieved by applying a learnable weight matrix \(\text{W}^u_\text{D} \in \mathbb{R}^{d \times d_D}\) and a bias vector \(b^u_\text{D} \in \mathbb{R}^{d}\), thus forming the final representation \(\text{H}_{D_{i}}\):

\begin{equation}
\text{H}_{D_{i}} = \text{W}^u_\text{D} \text{E}^{\text{enc}}_{D_{i}} + b^u_\text{D}, \quad  \text{H}_{D_i} \in \mathbb{R}^{d\times L_{D_i}}
\end{equation}

\subsection{Input, Process, Output Projection Heads}
\subsubsection{Text Artefact Originated Workflow Signals}
The encoded $\texttt{[CLS]}$ token representation of $T_i$, $h_{i, \texttt{[CLS]}_\text{T}} \in \text{H}_{T_i}$ is retrieved as the learned representative vector,  $h^{\texttt{[REP]}}_{i, \text{T}} = h_{i, \texttt{[CLS]}_\text{T}}$ and linearly projected by three separate Projection Heads: Input ($\text{W}^{\text{I}}_{\text{T}}$, $b^{\text{I}}_{\text{T}}$), Process ($\text{W}^{\text{P}}_{\text{T}}$, $b^{\text{P}}_{\text{T}}$) and Output ($\text{W}^{\text{O}}_{\text{T}}$, $b^{\text{O}}_{\text{T}}$), to obtain the following Workflow Signals:

\begin{align}
    i_{T_i} &= \text{W}^{\text{I}}_{\text{T}}h^{\texttt{[REP]}}_{i, \text{T}} + b^{\text{I}}_{\text{T}} \in \mathbb{R}^{d}\\
    p_{T_i} &= \text{W}^{\text{P}}_{\text{T}}h^{\texttt{[REP]}}_{i, \text{T}} + b^{\text{P}}_{\text{T}} \in \mathbb{R}^{d}\\
    o_{T_i} &= \text{W}^{\text{O}}_{\text{T}}h^{\texttt{[REP]}}_{i, \text{T}} + b^{\text{O}}_{\text{T}} \in \mathbb{R}^{d}
\end{align}

\subsubsection{Image Artefact Originated Workflow Signals}
We define $h^{\texttt{[REP]}}_{i, \text{F}} = \text{MaxPooling}(\text{H}_{F_i})$.
$h^{\texttt{[REP]}}_{i, \text{F}}$ is linearly projected by three separate Projection Heads: Input ($\text{W}^{\text{I}}_{\text{F}}$, $b^{\text{I}}_{\text{F}}$), Process ($\text{W}^{\text{P}}_{\text{F}}$, $b^{\text{P}}_{\text{F}}$) and Output ($\text{W}^{\text{O}}_{\text{F}}$, $b^{\text{O}}_{\text{F}}$), to obtain the following Workflow Signals:

\begin{align}
    i_{F_i} &= \text{W}^{\text{I}}_{\text{F}}h^{\texttt{[REP]}}_{i, \text{F}} + b^{\text{I}}_{\text{F}} \in \mathbb{R}^{d}\\
    p_{F_i} &= \text{W}^{\text{P}}_{\text{F}}h^{\texttt{[REP]}}_{i, \text{F}} + b^{\text{P}}_{\text{F}} \in \mathbb{R}^{d}\\
    o_{F_i} &= \text{W}^{\text{O}}_{\text{F}}h^{\texttt{[REP]}}_{i, \text{F}} + b^{\text{O}}_{\text{F}} \in \mathbb{R}^{d}
\end{align}

\subsubsection{Document Artefact Originated Workflow Signals}
The encoded $\texttt{[CLS]}_{\text{T}}$ representations of each text elements and the MaxPooled representations of each image elements of $D_i$, $\{h^{\texttt{[REP]}}_{q, _\text{F$\vee$T} }\}_q \in \text{H}_{D_i}$ are retrieved and averaged into $h^{\texttt{[REP]}}_{i, \text{D}} \in \mathbb{R}^{d}$. The resulting vector is linearly projected by three separate Projection Heads: Input ($\text{W}^{\text{I}}_{\text{D}}$, $b^{\text{I}}_{\text{D}}$), Process ($\text{W}^{\text{P}}_{\text{D}}$, $b^{\text{P}}_{\text{D}}$) and Output ($\text{W}^{\text{O}}_{\text{D}}$, $b^{\text{O}}_{\text{D}}$), to obtain the following Workflow Signals:

\begin{align}
    i_{D_i} &= \text{W}^{\text{I}}_{\text{D}}h^{\texttt{[REP]}}_{i, \text{D}} + b^{\text{I}}_{\text{D}} \in \mathbb{R}^{d}\\
    p_{D_i} &= \text{W}^{\text{P}}_{\text{D}}h^{\texttt{[REP]}}_{i, \text{D}} + b^{\text{P}}_{\text{D}} \in \mathbb{R}^{d}\\
    o_{D_i} &= \text{W}^{\text{O}}_{\text{D}}h^{\texttt{[REP]}}_{i, \text{D}} + b^{\text{O}}_{\text{D}} \in \mathbb{R}^{d}
\end{align}

\section{Decoding Intention}
We define a Workflow Intention $\gamma$ as a triple of Input, Process and Output Workflow Signals:

\begin{equation}
\gamma = (i_{\gamma}, p_{\gamma}, o_{\gamma})
\end{equation}

We define the Workflow Intention Set of a set of Business Artefacts $\mathcal{A}$ as a set of Workflow Intentions $\Gamma = \{\gamma_i \}_i$. The goal is to generate the Workflow Intention Set, i.e. Workflow Intention object(s), from a contextual representation of all the Business Artefacts. To do so we employ an encoder-decoder architecture described as follows.

\subsection{Intra-Modality Attention}
Across multiple Business Artefacts $\mathcal{A}_{m_k} = \{a_{m_k, i}\}_i$ of the same modality $m_k$, the encoded sequences $\{\text{E}^{\text{enc}}_{m_k, i}\}_i$ are concatenated, encoded by the self-attention based encoder $\textbf{Encoder}^{\text{intra}}_{m_k}$ into $\text{H}^{\text{intra}}_{\mathcal{A}_{m_k}}$. 
An encoded $\texttt{[REP]}$ token representation $h^{\text{intra}, \texttt{[REP]}}_{\mathcal{A}_{m_k}}$ is computed, linearly projected into the unified space by $(\text{W}^{\text{intra},u}_{m_k}$, $b^{\text{intra}, u}_{m_k})$ then by three Projection Heads: Input ($\text{W}^{\text{intra}, \text{I}}_{m_k}$, $b^{\text{intra}, \text{I}}_{m_k}$), Process ($\text{W}^{\text{intra}, \text{P}}_{m_k}$, $b^{\text{intra}, \text{P}}_{m_k}$) and Output ($\text{W}^{\text{intra}, \text{O}}_{m_k}$, $b^{\text{intra}, \text{O}}_{m_k}$).

\subsubsection{Artefact vectors Aggregation}
For a given modality \(m_k \in \{\text{T}, \text{F}, \text{D}\}\), let $\mathcal{A}_{m_k} = \{a_{m_k, i}\}_i$ be a set of Business Artefacts of this modality and $\{\text{E}^{\text{enc}}_{m_k, i}\}_i$ be the set of encoded tensors of these Business Artefacts.

$\forall i, \text{E}^{\text{enc}}_{m_k, i} \in \mathbb{R}^{d_{m_k} \times L_{m_k, i}}$ where $L_{m_k, i}$ is the number of encoded vectors of $a_{m_k, i}$.

\begin{equation}
    \text{E}_{\mathcal{A}_{m_k}} = \text{Concat}(\{\text{E}^{\text{enc}}_{m_k, i}\}_i) \in \mathbb{R}^{d_{m_k} \times L_{\mathcal{A}_{m_k}}} \text{ where } L_{\mathcal{A}_{m_k}} = \sum_{i=1}^{|\mathcal{A}_{m_k}|}L_{m_k, i}
    \label{eq:artefact_modality_agg_matrice}
\end{equation}

\subsubsection{Intra-Modality Encoder and Signals}
To capture contextual dependencies, we pass $\text{E}_{\mathcal{A}_{m_k}}$ through the encoder  $\textbf{Encoder}^{\text{intra}}_{m_k}$ to calculate  $\text{E}^{\text{intra}}_{\mathcal{A}_{m_k}}$ which is linearly projected by $(\text{W}^{\text{intra}, u}_{m_k}$, $b^{\text{intra}, u}_{m_k})$ to obtain $\text{H}^{\text{intra}}_{\mathcal{A}_{m_k}} \in \mathbb{R}^{d \times L_{\mathcal{A}_{m_k}}}$.\\

The representative encoded $\texttt{[REP]}$ token representation of $\mathcal{A}_{m_k}$ is computed as \\ $h^{\text{intra}, \texttt{[REP]}}_{\mathcal{A}_{m_k}} =  \text{MaxPooling}(\text{H}^{\text{intra}}_{\mathcal{A}_{m_k}})$ and linearly projected by the three separate Projection Heads: Input ($\text{W}^{\text{intra}, \text{I}}_{m_k}$, $b^{\text{intra}, \text{I}}_{m_k}$), Process ($\text{W}^{\text{intra}, \text{P}}_{m_k}$, $b^{\text{intra}, \text{P}}_{m_k}$) and Output ($\text{W}^{\text{intra}, \text{O}}_{m_k}$, $b^{\text{intra}, \text{O}}_{m_k}$), to obtain the following Workflow Signals:

\begin{align}
    i_{\mathcal{A}_{m_k}} &= \text{W}^{\text{intra}, \text{I}}_{m_k}h^{\text{intra}, \texttt{[REP]}}_{\mathcal{A}_{m_k}} + b^{\text{intra}, \text{I}}_{m_k} \in \mathbb{R}^{d}\\
    p_{\mathcal{A}_{m_k}} &= \text{W}^{\text{intra}, \text{P}}_{m_k}h^{\text{intra}, \texttt{[REP]}}_{\mathcal{A}_{m_k}} + b^{\text{intra}, \text{P}}_{m_k} \in \mathbb{R}^{d}\\
    o_{\mathcal{A}_{m_k}} &= \text{W}^{\text{intra}, \text{O}}_{m_k}h^{\text{intra}, \texttt{[REP]}}_{\mathcal{A}_{m_k}} + b^{\text{intra}, \text{O}}_{m_k} \in \mathbb{R}^{d}
\end{align}

\subsection{Inter-Modality Fusion Attention}
Considering  $\mathcal{A} = \{\mathcal{A}_{m_k}\}_k$ a set of  Business Artefacts grouped by modality. From now on, we consider $\mathcal{A}_{\text{T}}$,  $\mathcal{A}_{\text{F}}$, $\mathcal{A}_{\text{D}}$ sets of text, image and document Business Artefacts respectively.

\subsubsection{Inter-Modality vectors Aggregation}
We form a combined matrix $\text{H}^{\text{inter}}$ by concatenating the intra-modality encoder outputs $\{\text{H}^{\text{intra}}_{\mathcal{A}_{m_k}}\}_k$ column-wise:

\begin{equation}
    \text{H}^{\text{inter}} = \text{Concat}(\{\text{H}^{\text{intra}}_{\mathcal{A}_{m_k}}\}_k) \in \mathbb{R}^{d \times L_{\mathcal{A}}} \text{ where } L_{\mathcal{A}} = \sum_{k=1}^{|\mathcal{A}|}L_{\mathcal{A}_{m_k}}
    \label{eq:inter_agg_matrice}
\end{equation}

\subsubsection{Fusion Encoder}
We pass $\text{H}^{\text{inter}}$ through an encoder network $\textbf{Encoder}_{\text{fusion}}$ to calculate $\text{H}^{\text{fusion}} \in \mathbb{R}^{d \times L_{\mathcal{A}}}$. We currently do not employ projection heads to compute the Workflow Signals out of the fused representation of the Business Artefacts, as the fusion encoder is trained on Workflow Intention generation and not Workflow Signal extraction, as described later in the paper.

\subsection{Intention Decoder}
The decoder generates vectors based on the context computed from the artefacts. Each generated vector is projected into Workflow Signals $i_{\gamma}, o_{\gamma}$ and $ p_{\gamma}$, defining a Workflow Intention object $\gamma$ as an element of the Workflow Intention Set $\Gamma$. It is made of $N_{\text{decoder}}$ layers. Each layer is composed of a block of $n_{\text{decoder}}$ masked self-attention heads coupled with a LayerNorm block, followed by a block of $n_{\text{decoder}}$ cross-attention heads coupled with a LayerNorm block.

\subsubsection{Generation loop}
We initialize a decoded sequence $\text{S}^{\text{dec}}_0$ with a $\texttt{[BOS]}$ token embedding representation $\textbf{E}_{\text{fusion}}(\texttt{[BOS]})$.

At iteration $t$, in each decoder layer, $\text{S}^{\text{dec}}_t$ is first encoded through the masked multi-head self-attention heads, then attends to the fusion encoder's multimodal Business Artefact context $\text{H}^{\text{fusion}}$ via the cross-attention heads. The output sequence encoded by all the layers is denoted as $\tilde{\text{S}}^{\text{dec}}_t$. The last vector of the sequence, denoted by 
$\tilde{\text{s}}^{\text{dec}}_{t, -1} \in \mathbb{R}^{d}$ is linearly projected by ($\text{W}_{\gamma}$, $b_{\gamma}$) to produce $\tilde{\gamma}_t \in \mathbb{R}^{d}$.
We introduce two stopping mechanisms below: the Stopping Head and the Stopping Criteria. The Stopping Head acts as a first layer to stop the generation based the latest computed context. The Stopping Criteria stops the generation based on the latest generated Workflow Intention object. 

If the Stopping Head described below suggests to accept the generation, $\tilde{\gamma}_t \in \mathbb{R}^{d}$ is linearly projected by three separate Projection Heads: Input ($\text{W}^{\text{I}}_{\gamma}$, $b^{\text{I}}_{\gamma}$), Process ($\text{W}^{\text{P}}_{\gamma}$, $b^{\text{P}}_{\gamma}$) and Output ($\text{W}^{\text{O}}_{\gamma}$, $b^{\text{O}}_{\gamma}$) to obtain the following Workflow Signals:

\begin{align}
    i_{\gamma_t} &= \text{W}^{\text{I}}_{\gamma}\tilde{\gamma}_t + b^{\text{I}}_{\gamma} \in \mathbb{R}^{d}\\
    p_{\gamma_t} &= \text{W}^{\text{P}}_{\gamma}\tilde{\gamma}_t + b^{\text{P}}_{\gamma} \in \mathbb{R}^{d}\\
    o_{\gamma_t} &= \text{W}^{\text{O}}_{\gamma}\tilde{\gamma}_t + b^{\text{O}}_{\gamma} \in \mathbb{R}^{d}
\end{align}

These projections produce the Intention object:

\begin{equation}
\gamma_t = (i_{\gamma_t}, p_{\gamma_t}, o_{\gamma_t})
\end{equation}
If the Stopping Criteria described below suggest to accept and continue the generation, we start iteration $t+1$ with: 

\begin{equation}
    \text{S}^{\text{dec}}_{t+1} = \text{Concat}(\text{S}^{\text{dec}}_{t}, \tilde{\gamma_t})
\end{equation}

Let $t_f$ be the last iteration that passed the Stopping Mechanisms, we have:

\begin{equation}
    \Gamma = \{\gamma_t \}_{t=1}^{t_f}
\end{equation}

\subsubsection{Stopping Mechanisms}
\paragraph{Stopping Head}

We define the Stopping Head as 
\[
\text{MLP}_{\text{stop}} = \text{MLP}(\text{ReLU}, n_{\text{stop}}, (\text{W}_{\text{stop}, i}, \text{b}_{\text{stop}, i})_{i=1}^{n_{\text{stop}}}, (0, 1))
\]
 where 0 denotes the ``Stop" class to stop the generation and 1 the ``Accept" class to accept the current generation a priori.
The intuition is to decide if the current generated sequence of Workflow Intentions, attended with the Business Artefacts context, is complete or not.

\begin{align}
    &\forall t>1, \text{MLP}_{\text{stop}}(\tilde{\text{s}}^{\text{dec}}_{t, -1}) = \delta_t^{\text{head}} \in \{0, 1\} \\
    &\text{With } \delta_t^{\text{head}} = \begin{cases}
        1 & \text{if } \mathbb{P}_t(\text{``Accept"}) > 0.5 \\
        0 & \text{else}
    \end{cases} \text{ and } \delta^{\text{head}}_1 = 1
\end{align}

\paragraph{Stopping Criteria}
We define the Redundant Stopping Criterion as
\begin{align}
    &\forall t>1, \delta_t^{\text{sim}} = \begin{cases} 
        1 & \text{if }  \frac{1}{3}\big(
        \frac{<i_{\gamma_{t'}}, i_{\gamma_{t}}>}{\|i_{\gamma_{t'}}\|\|i_{\gamma_{t}}\|} + 
        \frac{<p_{\gamma_{t'}}, p_{\gamma_{t}}>}{\|p_{\gamma_{t'}}\|\|p_{\gamma_{t}}\|} + 
        \frac{<o_{\gamma_{t'}}, o_{\gamma_{t}}>}{\|o_{\gamma_{t'}}\|\|o_{\gamma_{t}}\|}
        \big) < \tau^{\text{sim}} \quad \forall t' \in \llbracket1, t-1\rrbracket\\
        0 &\text{else}
        \end{cases} \\
    &\text{With } \delta_1^{\text{sim}} = 1 \text{ and } \tau^{\text{sim}} \in [0,1]
\end{align}

At step $t$, $\delta_t^{\text{sim}} = 1$ indicates to continue the generation whereas $\delta_t^{\text{sim}} = 0$ indicates to stop the generation. The intuition is to refuse $\gamma_t$ and stop the generation if the generated Workflow Intention at step $t$ is too similar to one of the previously generated Workflow Intention.\\

We define the Hard Stopping Criterion by $t_{\text{max}}$ such that if $t > t_{\text{max}}$ the generation is stopped. This means that we constrain a user query to not include more than $t_{\text{max}}$ distinct Workflow Intentions.

\section{Training}
\subsection{Phase 1: Business Artefacts Encoding and Signal Extraction}
We employ a two stage training regimen. First, we train each modality independently for each Business Artefact in each modality $m_k$ where we have an $\textbf{Encoder}_{m_k}$ that is finetuned and $(\text{W}^u_{m_k}$, $b^u_{m_k}), (\text{W}^\text{I}_{m_k}$, $b^\text{I}_{m_k}),  (\text{W}^\text{P}_{m_k}$, $b^\text{P}_{m_k}), (\text{W}^\text{O}_{m_k}$, $b^\text{O}_{m_k}$) are trained by passing $N^{(1.1)}_{\text{artefact}, m_k}$ Business Artefacts for each modality. \\
For all modalities $m_k, \mathcal{A}^{(1.1)}_{m_k} = \{a^{(1.1)}_{m_k, i}\}_{i=1}^{N^{(1.1)}_{\text{artefact}, m_k}}$ denotes the set of training Business Artefacts for modality $m_k$ at Stage 1 of Phase 1 ($N^{(1.1)}_{\text{artefact}, m_k} = |\mathcal{A}^{(1.1)}_{m_k}|$). \\ 

Then, we continue training each modality independently over the intra-modality layers so that 
for all $m_k$,  $\textbf{Encoder}_{m_k}^{\text{intra}}$ is finetuned and $(\text{W}^{\text{intra}, u}_{m_k}, b^{\text{intra}, u}_{m_k})$, $(\text{W}^{\text{intra}, \text{I}}_{m_k}, b^{\text{I}, \text{intra}}_{m_k})$, $(\text{W}^{\text{intra}, \text{P}}_{m_k}, b^{\text{intra}, \text{P}}_{m_k})$, $(\text{W}^{\text{intra}, \text{O}}_{m_k}, b^{\text{intra}, \text{O}}_{m_k})$ are trained by passing $N^{(1.2)}_{\text{set}, m_k}$ Business Artefact sets for each modality.\\
For all modalities $m_k$, we provide $\{\mathcal{A}^{(1.2)}_{m_k, j}\}_{j=1}^{N^{(1.2)}_{\text{set}, m_k}} = \{\{a^{(1.2)}_{m_k, i,j}\}_{i=1}^{|\mathcal{A}^{(1.2)}_{m_k, j}|}\}_{j=1}^{N^{(1.2)}_{\text{set}, m_k}}$ denoting the sets of training Business Artefacts for modality $m_k$ at Stage 2 of Phase 1.\\

In total, $\sum_{m_k}N^{(1.1)}_{\text{artefact}, m_k}$ Business Artefacts are provided in stage 1 and  $\sum{_{m_k}}\sum_{j=1}^{N^{(1.2)}_{\text{set}, m_k}}|\mathcal{A}^{(1.2)}_{m_k, j}|$ in stage 2.

\subsubsection{Classification Tasks for i, o and p}
We build ground truth data based on three sets of text elements $\text{I}_g, \text{P}_g$ and $\text{O}_g$:
\begin{itemize}
\item $\text{I}_g$: elements that serve as input Workflow Signals within the Business Artefacts.
\item $\text{P}_g$: elements that relate to transformations or Processes within the Business Artefacts.
\item $\text{O}_g$: elements that describe expected output Workflow Signals within the Business Artefacts.
\end{itemize}

Each projected vector $i, p$ and $o$ is associated with a ground truth representation over its set denoted by:
\begin{equation}
\text{C}_x^* \in \mathbb{R}^{|\text{X}_g| \times (M+2)}, \quad \text{where } (x, \text{X}) \in \{(i, \text{I}), (p, \text{P}), (o, \text{O})\},  M \in \mathbb{N}^*
\end{equation}

Each projected vector $x \in \{i, o, p\}$ is passed through a dedicated $\text{MLP}_\text{X}$ to predict discrete counts for each class in $\text{X}_g$ up to a maximum count $M$.\\

Each classifier ($\text{MLP}_\text{I}$, $\text{MLP}_\text{P}$, $\text{MLP}_\text{O}$) outputs a set of logits for each class:
\begin{equation}
\hat{\text{C}}_x \in \mathbb{R}^{|\text{X}_g| \times (M+2)}, \quad \text{where } (x, \text{X}) \in \{(i, \text{I}), (p, \text{P}), (o, \text{O})\}
\end{equation}

Each row i of $\hat{\text{C}}_x $ represents the unnormalized logits for predicting the count class $c$ of the corresponding element $x_{g, i}$ of $\text{X}_g$ where:
\begin{itemize}
    \item c=0: $x_{g, i}$ not present
    \item $c\in \llbracket1, M\rrbracket$: $x_{g, i}$ referenced in plural form with known exact count $c$.
    \item $c = M+1$: $x_{g, i}$ referenced in plural form, but exact count is unknown.
    
\end{itemize}

\subsubsection{Loss}
For each class indexed by $k \in \llbracket1, |\text{X}_g|\rrbracket$, given a ground-truth count class $c^*_{k} \in \llbracket0, M+1\rrbracket$, we apply a categorical cross-entropy loss:
\begin{equation}
\mathcal{L}_{\text{X}, k} = -\sum_{m=0}^{M+1} \delta_m(c^*_{k}) \log \big(\text{softmax}(\hat{\text{C}}_x[k])[m]\big) \text{ where }  \delta_m: x \mapsto \begin{cases}
    1 &\text{if } x = m \\
    0 & \text{else}
\end{cases} 
\end{equation}

The total loss for each head is computed as:
\begin{equation}
\mathcal{L_\text{X}} = \frac{1}{|\text{X}_g|} \sum_{k=1}^{|\text{X}_g|} \mathcal{L}_{\text{X},k}
\end{equation}

The overall loss is the sum over the three heads:
\begin{equation}
\label{eq:LSignal}
\mathcal{L}_{\text{signal}} = \text{bound}(\mathcal{L}_{\text{I}} + \mathcal{L}_{\text{P}} + \mathcal{L}_\text{O}, \lambda, \mu)
\end{equation}

With the bounding function $\text{bound}(\mathcal{L}, \lambda, \mu) = \frac{1}{1 + e^{-\lambda(\mathcal{L} - \mu)}}, \mu \in [0, 1]$ and $\lambda > 0$.

\paragraph{Stage 1}
$\forall m_k, \mathcal{A}^{(1.1)}_{m_k} = \{a^{(1.1)}_{m_k, i}\}_{i=1}^{N^{(1.1)}_{\text{artefact}, m_k}}$ denotes the set of training Business Artefacts for modality $m_k$ at Stage 1 of Phase 1. Each Business Artefact is tokenized, encoded, projected then classified.

\paragraph{Stage 2}
At this stage, all the elements of stage 1 are frozen.\\

$\forall m_k, \{\mathcal{A}^{(1.2)}_{m_k, j}\}_{j=1}^{N^{(1.2)}_{\text{set}, m_k}} = \{\{a^{(1.2)}_{m_k, i, j}\}_{i=1}^{|\mathcal{A}^{(1.2)}_{m_k, j}|}\}_{j=1}^{N^{(1.2)}_{\text{set}, m_k}}$ denotes the sets of training Business Artefacts for modality $m_k$ at Stage 2 of Phase 1. Each Business Artefact of each set is tokenized and encoded, then across each set the encoded Business Artefacts are concatenated, encoded, projected then classified. We define the ground truth over an Business Artefact set as $\text{C}_{x, {\mathcal{A}^{(1.2)}_{m_k, j}}}^*, \forall m_k, \forall j \in \llbracket1, N^{(1.2)}_{\text{set}, m_k}\rrbracket$.

\subsection{Phase 2: Decoding Intention}
In this phase, all the elements from Phase 1 are frozen.
The training data for this phase is $\mathcal{A}^{(2.2)} = \{\mathcal{A}^{(2.2)}_{q}\}_{q=1}^{N^{(2.2)}}$ where $N^{(2.2)}$ denotes the number of samples.
Each sample is such that $\mathcal{A}^{(2.2)}_{q} = \{\mathcal{A}^{(2.2)}_{q, m_k}\}_{m_k}$ where $\mathcal{A}^{(2.2)}_{q, m_k}$ is a set of Business Artefacts of modality $m_k$. For each sample, across each modality, across each set, each Business Artefact is tokenized and encoded. Across each set, encoded Business Artefacts are concatenated and encoded by the intra-modality encoder and projected in the $d$ dimension. Across each modality, the intra-modality encoded vectors are concatenated and the resulting sequence of vectors by $\textbf{Encoder}_{\text{fusion}}$. The decoding loop starts, provided with the entire context of the sample $\mathcal{A}^{(2.2)}_{q}$ from $\textbf{Encoder}_{\text{fusion}}$. $\textbf{Decoder}_{\text{Intention}}$ is producing a sequence $\hat{\Gamma} = \{\hat{\gamma_t}\}_{t=1}^{|\hat{\Gamma}|}$. We train the system by classifying $\hat{i}_{\gamma_t}, \hat{p}_{\gamma_t}, \hat{o}_{\gamma_t}$
of each $\gamma_t$ using $\text{MLP}^{\gamma}_{\text{I}}, \text{MLP}^{\gamma}_{\text{O}}, \text{MLP}^{\gamma}_{\text{P}}$ and compute the loss (described below) over a ground truth $i^*_{\gamma_t}, p^*_{\gamma_t}$ and $o^*_{\gamma_t}$ expressed over the sets $\text{I}_g, \text{P}_g, \text{O}_g$ as done for Stage 1.

\paragraph{Stopping Head and Intention Generation Losses}
Given a generated Workflow Intention: $\hat{\Gamma} = \{\hat{\gamma}_t \}_{t=1}^{\hat{t}_f}$ and a ground truth $\Gamma^* = \{\gamma_t^* \}_{t=1}^{t^*_f}$: 

\begin{equation}
    \forall \gamma \in \hat{\Gamma} \cup \Gamma^*,
\gamma =(i_{\gamma}, p_{\gamma}, o_{\gamma})
\end{equation}

We use $\mathcal{L}_{\text{signal}}$ defined previously and introduce a threshold $\tau_{\gamma}$ to consider two Workflow Intentions $\gamma_1, \gamma_2$ matching if and only if $\mathcal{L}_{\text{signal}}(\gamma_1, \gamma_2) \leq \tau_{\gamma}$.\\

We introduce the Coverage measure between $\Gamma^*$ and $\hat{\Gamma}$ as:
\begin{align}
    &\text{Coverage}_{\Gamma}(\Gamma^*, \hat{\Gamma}) = \frac{1}{t^*_f}
    \sum_{t=1}^{t^*_f} c_{\gamma,t} \\
    &\text{where } c_{\gamma,t} = \begin{cases}
        1 &\text{if } \min\limits_{\gamma \in \hat{\Gamma}}(\{{\mathcal{L}_{\text{signal}}(\gamma_t^*, \gamma)}\}) < \tau_{\gamma} \\
        0 & \text{else}
    \end{cases}
\end{align}

We define
\begin{itemize}
    \item For overlength: $\Delta_\Gamma^+ = \max(0,\, \hat{t}_f - t^*_f)$
    \item For underlength: $\Delta_\Gamma^-= \max(0,\, t^*_f - \hat{t}_f)$
\end{itemize}

We define the Workflow Intention sequence loss in terms of coverage, overlength and underlength, such as
\begin{align}
    \label{eq:LsequenceIntention}
   & \mathcal{L}_{\text{sequence}} = 1 - \Biggr[ \alpha_{\Gamma,c} \cdot 
    \text{Coverage}_{\Gamma}
    \;+\;
    \alpha_{\Gamma,o}\ \frac{1}{1 + \Delta_\Gamma^+}
    \;+\;
    \alpha_{\Gamma,u}\  \frac{1}{1 + \Delta_\Gamma^-} \Biggl] \notag\\
    &\text{where } \alpha_{\Gamma,c} + \alpha_{\Gamma,o} + \alpha_{\Gamma,u} = 1,\\
    &0\leq\alpha_{\Gamma,c}\leq1, \text{ } 0\leq\alpha_{\Gamma,o}\leq1 \text{ and }0\leq\alpha_{\Gamma,u}\leq1 \notag
    \
\end{align}

We define the Workflow Intention contrastive loss to encourage diverse Workflow Intention generation as:
\begin{align}
   \mathcal{L}_{\text{contrastive}} = \begin{cases}
    \frac{2}{\hat{t}_f(\hat{t}_f-1)}
   \sum\limits_{m=1}^{\hat{t}_f}
   \sum\limits_{n=m+1}^{\hat{t}_f} e^{-(\| i_{\gamma_m} - i_{\gamma_n} \|^2 + \| p_{\gamma_m} - p_{\gamma_n} \|^2 + \| o_{\gamma_m} - o_{\gamma_n} \|^2)}  &\text{ if }\hat{t}_f > 1\\
   0 &\text{ otherwise}
   \end{cases}
\end{align}
\noindent
with $\forall i, \gamma_i \in \hat{\Gamma}$. \\

We define the Stopping Head loss as:

\begin{align}
    &\mathcal{L}_{\text{head}} = -\frac{1}{\text{max}(\hat{t}_f,t^*_f)} \sum_{t=1}^{\text{max}(\hat{t}_f,t^*_f)}  \left[ \delta^{*\text{head}}_t \log (\mathbb{P}_t(\text{``Accept"})) + (1 - \delta^{*\text{head}}_t) \log (1 - \mathbb{P}_t(\text{``Accept"})) \right] \\
    &\text{with } \begin{cases}
        \mathbb{P}_t(\text{``Accept"}) = 0  &\forall t > t^*_f \text{ if } t^*_f > \hat{t}_f \\
        \delta^{*\text{head}}_t = 0  &\forall t > t^*_f \text{ if } t^*_f < \hat{t}_f \\
    \end{cases} \notag
\end{align}

We define the Workflow Intention Loss as:

\begin{equation}
    \mathcal{L}_{\text{Intention}} = \mathcal{L}_{\text{head}} + \mathcal{L}_{\text{contrastive}} + \mathcal{L}_{\text{sequence}}
\end{equation}

\vspace{1em}

\section{Computational Complexity}

Parameter count analysis reveals significant overhead:
\begin{itemize}
    \item Text encoder: $24 \text{ layers} \times [(16 \text{ heads} \times 3 \text{ matrices} \times 64 \times 1024) + (1024 \times 1024) + (1024 \times 4096 \times 2)] \approx 300\text{M parameters}$
    \item Image encoder: $588\times3200 + 45 \text{ layers} \times [(25 \text{ heads} \times 3 \text{ matrices} \times 128 \times 3200) + (3200 \times 3200) + (3200 \times 12800 \times 2)] \approx 5.5\text{B parameters}$
    \item Document encoder:
    \begin{itemize}
        \item Text Encoder $\approx$ 300M
        \item Image Encoder $\approx$ 5.5B
        \item Document Encoder same as Text Encoder $\approx$ 300M
    \end{itemize}
    $\approx$ 6B parameters
    \item Fusion Encoder: $24 \text{ layers} \times [(128 \text{ heads} \times 3 \text{ matrices} \times 8 \times 1024) + (1024 \times 1024)+ (1024 \times 65536 \times 2)] \approx 3.3\text{B parameters}$
    \item Intention Decoder: $24 \text{ layers} \times [2\times((128 \text{ heads} \times 3 \text{ matrices} \times 8 \times 1024) + (1024 \times 1024)) + (1024 \times 65536 \times 2)] \approx 3.5\text{B parameters}$
    \
    \item Projection Heads: 
    \begin{itemize}
        \item Text Unifier: $ 1024 \times 1024$
        \item Text Input, Output, Process Projections: $ 3\text{ heads} \times 1024 \times 1024$
        \item Image Unifier: $ 3200 \times 1024$
        \item Image Input, Output, Process Projections: $ 3\text{ heads} \times 1024 \times 1024$
        \item Document Unifier: $3200 \times 1024 + 1024 \times 1024$
        \item Document Input, Output, Process Projections: $3\text{ heads} \times 1024 \times 1024$
        \item Decoder: $1024 \times 1024$
        \item Decoder Input, Output, Process: $3\text{ heads} \times 1024 \times 1024$
    \end{itemize}
    $2 \text{ [per Artefact and Intra modality]} \times [
            1024 \times 1024 +
            3 \times 1024 \times 1024 +
            3200 \times 1024 +
            3 \times 1024 \times 1024 +
            1024 \times 1024 +
            3 \times 1024 \times 1024
    ] + [3200 \times 1024] + [1024 \times 1024 + 3 \times 1024 \times 1024] \approx 38\text{M parameters}$    
    
    \item MLPs: Assuming $|\text{X}_g| \approx 10^5,
    \text{X}\in\{\text{I}, \text{P}, \text{O}\}$,
    \begin{itemize}
        \item $2 \text{ Signal Classifications [per Artefact and Intra modality]} \times 3 \text{ modalities}$
        \item $1 \text{ Attention Signal Classification}$
        \item $1 \text{ Stopping Mechanism}$
    \end{itemize}
    $7 \times 3 \text{ heads} \times [4096 \times 1024 + 4096 \times 10^5] + [4096 \times 1024 + 4096 \times 2] \approx 8.7\text{B parameters}$ 
\end{itemize}

This results in an approximate total of $27.5$ billion parameters, excluding the tokenizer and token encoder parameters, which are provided out of the box. \\

There are challenges due to the computational complexity of the system. The Workflow Intention framework exhibits $\mathcal{O}(n^2d + nd^2)$ complexity in the attention mechanisms, where $n$ represents the sequence length and $d$ represents the embedding dimension. This quadratic scaling becomes problematic in the inter-modality fusion encoder, where $n = 1+\sum_{i=1}^{|\mathcal{A}|} L_{\mathcal{A}_{m_k}}$.\\

The document encoder's representational overhead is particularly significant, requiring $2(L_{D_{p,i}}^{\text{T}} + L_{D_{p,i}}^{\text{F}})$ vectors of dimension $d_{\text{T}}$ for each document page. To optimize computational efficiency, the sequence length \( n \) can be reduced using sparsification techniques like Longformer (sliding window attention, Beltagy et al. \cite{longformer}) or Linformer (low-rank approximation, Wang et al. \cite{linformer}). The hidden dimension \( d \) in intermediate layers can be decreased to lower the \( \mathcal{O}(d^2) \) cost in the encoders. Additionally, compressing document representations via pre-processing pipelines would reduce the number of stored vectors per page and improve memory efficiency while preserving essential information. The computational complexities incurred by the length of the decoded sequence are negligible as the number of Intentions is typically bounded between 1 and 5.

\section{Conclusion}
In this paper, we have introduced Workflow Intention, a comprehensive framework for identifying and encoding Process objectives within complex business environments. Our approach addresses the fundamental challenge of interpreting and leveraging Process documentation through a systematic methodology that interprets and aligns Input, Process and Output Workflow Signals from diverse Business Artefacts. The mathematical framework we developed formalizes these Workflow Signals as vectors and Workflow Intentions as tensors, providing a rigorous foundation for understanding Process objectives.\\

The multimodal generative system we developed demonstrates the practical applicability of our framework, successfully processing various types of Business Artefacts through Modality-Specific Encoding, Intra-Modality Attention, Inter-Modality Fusion Attention and Intention Decoding. Our hierarchical encoder methodology effectively generates Workflow Intention from Workflow Signal across modalities. This work enables organizations to rapidly implement supervised automation and evolve legacy processes into efficient AI-enhanced Workflows enriched with best practice.

\bibliographystyle{plain}

\newpage

\appendix

\section{Appendix}
We describe the models and mechanisms we employ, including the backbone architectures and parametrization, as well as the mathematical interpretation of our framework.

\subsection{Classic Computational Mechanisms}

\subsubsection{softmax}
The softmax function is defined such that:
\begin{equation} 
    \forall \text{X}=(x_i)^{d_{\text{X}}}_{i=1} \in \mathbb{R}^{d_{\text{X}}}, 
    \text{softmax}(\text{X}) = \Big(\frac{e^{x_i}}{\sum_{j=1}^{d_{\text{X}}} e^{x_j}} \Big)_{i=1}^{d_{\text{X}}}  \in \mathbb{R}^{d_{\text{X}}}
\end{equation}

By default, for a matrix $\text{M} \in \mathbb{R}^{d\times n}$, $\text{softmax}(\text{X})$ denotes the row-wise application of the softmax function i.e. 

\begin{equation} 
    \text{softmax}(\text{M}) = (\text{softmax}(\text{M[i:]}))^{d}_{i=1}
\end{equation}

\subsubsection{LayerNorm}
The LayerNorm mechanism is described by a parameter $\epsilon$ and two learnable parameters $(\gamma, \beta) $ such that:
\begin{align} 
    &\forall \text{X}=(x_i)^{d_{\text{X}}}_{i=1} \in \mathbb{R}^{d_{\text{X}}}, \notag \\
    &\text{LayerNorm}(\text{X}) = \frac{X - \mu}{\sigma}.\gamma + \beta \\
    &\text{where } \mu = \frac{1}{d_{\text{X}}} \sum_{i=1}^{d_{\text{X}}} x_i, 
    \sigma_i = \sqrt{\frac{1}{d_{\text{X}}} \sum_{i=1}^{d_{\text{X}}} (x_i - \mu)^2 + \epsilon} \notag
\end{align}

\subsubsection{Linear Projection}
A linear projection mechanism $\text{L}$ is described by a learnable weight matrix and bias term $(\text{W}_{\text{L}}, \text{b}_{\text{L}})$ with $\text{W}_{\text{L}} \in \mathbb{R}^{d_{\text{L}}\times d_{\text{X}}}$, $\text{b}_{\text{L}} \in \mathbb{R}^{d_{\text{L}}}$ such that 
\begin{equation}
    \forall y \in \mathbb{N}^*, 
    \forall \text{X} \in \mathbb{R}^{d_{\text{X}}\times y}, 
    \text{L}(\text{X}) = \text{W}_{\text{L}}\text{X} + \text{b}_{\text{L}} \in \mathbb{R}^{d_{\text{L}}\times y}
\end{equation}
The bias is broadcast across all columns of $\text{W}_{\text{L}}\text{X}$.

\subsubsection{MLP}
A Multi-Layer-Perceptron $\text{MLP}$ is described by $n_{\text{MLP}}$ layers of Linear Projection and activation function:$((f_{\text{act},i},\text{W}_{i}, \text{b}_{i}))_{i=1}^{n_{\text{MLP}}}$, such that
\begin{align}
    &\forall y \in \mathbb{N}^*,
    \forall \text{X} \in \mathbb{R}^{d_{\text{X}}\times y}, \notag \\
    &\text{MLP}(\text{X}) = f_{\text{act}, n_{\text{MLP}}}\big(
    \text{W}_{n_{\text{MLP}}-1}
    f_{\text{act}, n_{\text{MLP}}-1}(...
    f_{\text{act}, 1} (\text{W}_1\text{X} +  \text{b}_1)
    ...)
    + \text{b}_{n_{\text{MLP}}}
    \big)
\end{align}

\paragraph{$\text{MLP}_{\text{I}}, \text{MLP}_{\text{O}}, \text{MLP}_{\text{P}}$ ; $\text{MLP}^{\text{intra}}_{\text{I}}, \text{MLP}^{\text{intra}}_{\text{O}}, \text{MLP}^{\text{intra}}_{\text{P}}$ ; $\text{MLP}^{\gamma}_{\text{I}}, \text{MLP}^{\gamma}_{\text{O}}, \text{MLP}^{\gamma}_{\text{P}}$; $\text{MLP}_{\text{stop}}$}
These MLP networks are such that $((\text{softmax},\text{W}^{\text{MLP}_\text{X}}_{1}, \text{b}^{\text{MLP}_\text{X}}_{1}), (\text{Id},\text{W}^{\text{MLP}_\text{X}}_{2}, \text{b}^{\text{MLP}_\text{X}}_2))$ with an inner dimension of 4096 i.e. $\text{W}^{\text{MLP}_\text{X}}_{1} \in \mathbb{R}^{4096\times 1024}$ and $\text{W}^{\text{MLP}_\text{X}}_{2} \in \mathbb{R}^{|\text{X}_g|\times 4096}$, $\text{X} \in \{\text{I}, \text{P}, \text{O}\}$.

\subsubsection{FFN}
A Feed-Forward-Network $\text{FFN}$ is described by $n_{\text{FFN}}$ layers of Linear Projection and activation function $((f_{\text{act},i},\text{W}_{i}, \text{b}_{i}))_{i=1}^{n_{\text{FFN}}}$, such that
\begin{align}
    &\forall y \in \mathbb{N}^*,
    \forall (\text{X}_i)_{i=1}^{y} \in \mathbb{R}^{d_{\text{X}}\times y}, \notag \\
    &\text{FFN}(\text{X}) = \Big(
    f_{\text{act}, n_{\text{FFN}}}\big(
    \text{W}_{n_{\text{FFN}}-1}
    f_{\text{act}, n_{\text{FFN}}-1}(...
    f_{\text{act}, 1} (\text{W}_1\text{X}_i +  \text{b}_1)
    ...)
    + \text{b}_{n_{\text{FFN}}}
    \big)
    \Big)_{i=1}^{y}
\end{align}

\subsection{Tokenizers and Token Encoders}
\label{sec:Tokenizer}

\paragraph{Text}
\label{sec:textTokenizer}
We employ the RoBERTa-large (Liu et al. \cite{roberta}) Byte-Level Byte Pair Encoding (BPE) tokenizer which has a 50 265 token vocabulary including the start-of-sequence token $\texttt{<s>}$ which we label as $\texttt{[CLS]}_\text{T}$ token. The token encoder embeds the tokens into $d_{\text{T}} = \mathbb{R}^{1024}$ as well as positional embeddings and sums both representations.

\paragraph{Image}
\label{sec:imageTokenizer}
We employ the tiling, unshuffling and flattening method of InternViT-6B-448px-V1.5 presented by Dosovitskiy et al. \cite{vit} and expended in InternVL by Chen et al. \cite{internVL}. Each image $F \in \mathbb{R}^{c\times h \times w}$ (with $c=3$) is resized in an optimal ratio $r^*(\tilde{t}) := w^*/h^*$ such that when divided into tiles $\{\tilde{t}_k\}_{k=1}^{N_{\tilde{t}}}$ where
$\forall k, \tilde{t}_k\in \mathbb{R}^{c\times \tilde{t} \times \tilde{t}}$ with $N_{\tilde{t}}$ = $h^* \times w^* / \tilde{t}^2$, $N_{\tilde{t}} < n_{\text{max}}$. For InternViT-6B-448px-V1.5, $\tilde{t}=448, n_{\text{max}}=12$. If $N_{\tilde{t}} > 1$, a thumbnail tile, which is a resized version of the image to the target dimension $\tilde{t}$, is added to the sequence of tiles. Each tile $\tilde{t}_k$ is then unshuffled 5 times by reducing the resolution dimensionality and increasing the number of channels, resulting in a sequence of patches $\{\tilde{p}_{k, i}\}_{i=1}^{N_{\tilde{p}}}$ where
$\forall i, \tilde{p}_{k, i}\in \mathbb{R}^{c\times \tilde{p} \times \tilde{p}}$ with $\tilde{p}$ = 14 and $N_{\tilde{p}} = 1024$ as $448\times448\times3 = (2^5\times2^5)\times(14\times14)\times 3 = 1024 \times(14\times14)\times 3 = 1024 \times 588$. Each of the $1024$ patches is flattened to $\mathbb{R}^{588}$ and linearly projected to $\mathbb{R}^{3200}$, thus $d_{\text{F}} = 3200$. The resulting sequence is added to learned positional embeddings from ViT (Dosovitskiy et al. \cite{vit}).

\paragraph{Document}
The Document tokenizer and token encoder process is fully described in the article, combining Text and Image tokenizers and token encoders.

\subsection{Attention Mechanisms}
\subsubsection{Self-Attention }
From \cite{attention}, given an input sequence represented as a matrix \( \text{X} \in \mathbb{R}^{d \times n} \), where \( d \) is the embedding dimension and \( n \) is the sequence length (number of tokens). Let $\text{W}^Q \in \mathbb{R}^{d_k \times d}, \text{W}^K \in \mathbb{R}^{d_k \times d}, \text{W}^V \in  \mathbb{R}^{d_v \times d}$ be learnable weight matrices. The Queries, Keys and Values are computed as $Q = \text{W}^Q \text{X} \in \mathbb{R}^{d_k \times n}, 
K = \text{W}^K \text{X} \in \mathbb{R}^{d_k \times n},
V = \text{W}^V \text{X} \in \mathbb{R}^{d_v \times n}
$.\\

The Attention is computed as 
\begin{equation}
    \text{Z} = V.\text{softmax}(A) \in  \mathbb{R}^{d_v \times n}, A = \frac{K^\top Q}{\sqrt{d_k}} \in \mathbb{R}^{n \times n}
\end{equation}

We denote $\text{Z} = \text{SelfAttention}(\text{X}, \text{W}^Q, \text{W}^K, \text{W}^V)$

\subsubsection{Masked Self-Attention}
Following previous notation, 

\begin{align}
    \text{Z} = V.\text{softmax}(A + M) \in  \mathbb{R}^{d_v \times n},
    M \in \mathbb{R}^{n \times n},
    M[i,j] =
    \begin{cases}
    0 & \text{if } i \geq j \\
    -\infty & \text{otherwise}
    \end{cases}
\end{align}

\subsubsection{Multi-Head Attention}
Let $H$ be the number of attention heads, $\text{W}^O$ a learnable weight matrix. The Multi-Head Attention is computed as
\begin{align}
    &\text{Z} = \text{W}^O.\text{Concat}(\{\text{SelfAttention}(\text{X},
    \text{W}_i^Q, \text{W}_i^K, \text{W}_i^V)\}_{i=1}^{H}) \in \mathbb{R}^{d_v \times n}, \\
    &\forall i, \text{W}_i^Q \in \mathbb{R}^{\frac{d_k}{H} \times d}, \text{W}_i^K \in \mathbb{R}^{\frac{d_k}{H} \times d}, \text{W}_i^V \in \mathbb{R}^{\frac{d_v}{H} \times d}, \text{W}^O \in \mathbb{R}^{d_v \times d_v} \notag
\end{align}

\subsubsection{Cross Attention}
Given an input sequence represented as a matrix \( \text{X} \in \mathbb{R}^{d \times n} \) and a context sequence represented as a matrix \( \text{Y} \in \mathbb{R}^{d \times m} \). Let $\text{W}^Q \in \mathbb{R}^{d_k \times d}, \text{W}^K \in \mathbb{R}^{d_k \times d}, \text{W}^V \in  \mathbb{R}^{d_v \times d}$ be learnable weight matrices. The Queries, Keys and Values are computed as $Q = \text{W}^Q \text{X} \in \mathbb{R}^{d_k \times n}, 
K = \text{W}^K \text{Y} \in \mathbb{R}^{d_k \times m},
V = \text{W}^V \text{Y} \in \mathbb{R}^{d_v \times m}
$.\\

The Attention is computed as 
\begin{equation}
    \text{Z} = V.\text{softmax}(A) \in  \mathbb{R}^{d_v \times n}, A = \frac{K^\top Q}{\sqrt{d_k}} \in \mathbb{R}^{m \times n}
\end{equation}

We denote $\text{Z} = \text{CrossAttention}(\text{X}, \text{Y}, \text{W}^Q, \text{W}^K, \text{W}^V)$

\subsection{Transformer Models}
\subsubsection{Encoder}
\label{sec:Encoder}
Each \textbf{Encoder} we employ uses Multi-Head Self-Attention and is described by $n_{\text{Enc}}$ layers and a dimension $d$.
Each layer $l$ is composed of:
\begin{enumerate}
    \item \textbf{Multi-Head Self-Attention:}
    \begin{itemize}
        \item $H_{\text{Enc}}$ Self-Attention heads, parametrized by $({\text{W}^{Q, l}_i}, {\text{W}^{K, l}_i}, {\text{W}^{V, l}_i})_{i=1}^{H_{\text{Enc}}}$, each matrix being in $\mathbb{R}^{d_{H_\text{Enc}}\times d}$ where $       d_{H_\text{Enc}} = d / H_\text{Enc}$
        \item A projection matrix $\text{W}^{O, l} \in \mathbb{R}^{d\times d}$
    \end{itemize}
    \item{\textbf{LayerNorm and Add:}} LayerNorm and the residual ``Add" operation are applied element-wise across each token's embedding dimension, i.e. column-wise when the input is structured as $d \times n$ where $d$ is the embedding dimension and $n$ is the sequence length.
    
    \item{\textbf{Feedforward Network (FFN):}}  $((\text{GELU},\text{W}^{\text{FFN}, l}_{1}, \text{b}^{\text{FFN}, l}_{1}), (\text{Id},\text{W}^{\text{FFN}, l}_{2}, \text{b}^{\text{FFN}, l}_{2}))$.

    \item{\textbf{LayerNorm and Add:}}
    LayerNorm and the residual ``Add" operation are applied element-wise across each token's embedding dimension, i.e. column-wise when the input is structured as $d \times n$ where $d$ is the embedding dimension and $n$ is the sequence length.

\end{enumerate}

\paragraph{Text $\textbf{Encoder}_{\text{T}}$}
\label{sec:textEncoder}
We employ the RoBERTa-large (Liu et al. \cite{roberta}) which has $n_{\text{Enc}} = 24$ and a final dimension $d_{\text{T}}=1024$. In each layer $l$, the Multi-Head Self-Attention is made of $H_{\text{Enc}} = 16$ heads. The Feedforward Network is defined as $((\text{GELU},\text{W}^{\text{FFN}, l}_{1}, \text{b}^{\text{FFN}, l}_{1}), (\text{Id},\text{W}^{\text{FFN}, l}_{2}, \text{b}^{\text{FFN}, l}_{2}))$ with an inner dimension of 4096 i.e. $\text{W}^{\text{FFN}, l}_{1} \in \mathbb{R}^{4096\times 1024}$ and $\text{W}^{\text{FFN}, l}_{2} \in \mathbb{R}^{1024\times 4096}$.

\paragraph{Image $\textbf{Encoder}_{\text{F}}$}
\label{sec:imageEncoder}
We employ the InternViT-6B-448px-V1-5 (Dosovitskiy et al., \cite{vit}) which has $n_{\text{Enc}} = 45$ and a final dimension $d_{\text{F}}=3200$. In each layer $l$, the Multi-Head Self-Attention is made of $H_{\text{Enc}} = 25$ heads. The Feedforward Network is defined as $((\text{GELU},\text{W}^{\text{FFN}, l}_{1}, \text{b}^{\text{FFN}, l}_{1}), \allowbreak (\text{Id},\text{W}^{\text{FFN}, l}_{2}, \text{b}^{\text{FFN}, l}_{2}))$ with an inner dimension of 12800 i.e. $\text{W}^{\text{FFN}, l}_{1} \in \mathbb{R}^{12800\times 3200}$ and $\text{W}^{\text{FFN}, l}_{2} \in \mathbb{R}^{3200\times 12800}$.

\paragraph{Document $\textbf{Encoder}_{\text{D}}$}
\label{sec:documentEncoder}
As for the text encoder, we employ the RoBERTa-large (Liu et al. \cite{roberta}) which has $n_{\text{Enc}} = 24$ and a final dimension $d_{\text{T}}=1024$. In each layer $l$, the Multi-Head Self-Attention is made of $H_{\text{Enc}} = 16$ heads. The Feedforward Network is defined as   $((\text{GELU},\text{W}^{\text{FFN}, l}_{1}, \text{b}^{\text{FFN}, l}_{1}), \allowbreak (\text{Id},\text{W}^{\text{FFN}, l}_{2}, \text{b}^{\text{FFN}, l}_{2}))$ with an inner dimension of 4096 i.e. $\text{W}^{\text{FFN}, l}_{1} \in \mathbb{R}^{4096\times 1024}$ and $\text{W}^{\text{FFN}, l}_{2} \in \mathbb{R}^{1024\times 4096}$. \paragraph{$\textbf{Encoder}^{\text{intra}}_{\text{T}}$, $\textbf{Encoder}^{\text{intra}}_{\text{F}}$, $\textbf{Encoder}^{\text{intra}}_{\text{D}}$}

These encoders are $\textbf{Encoder}_{\text{T}}, \textbf{Encoder}_{\text{F}}, \\
\textbf{Encoder}_{\text{D}}$ (trained in Phase 1 Stage 1), further trained in Phase 1 Stage 2.

\subsubsection{Decoder}
\label{sec:Decoder}
The \textbf{Decoder} we employ with an \textbf{Encoder} uses Masked Multi-Head Self-Attention and Multi-Head Cross Attention described by $n_{\text{Dec}}$ layers and a dimension $d$.
Each layer $l$ is composed of:
\begin{enumerate}
    \item \textbf{Masked Multi-Head Self-Attention:}
    \begin{itemize}
        \item $H^{\text{self}}_{\text{Dec}}$ Self-Attention heads, parametrized by $({\text{W}^{Q,\text{self}, l}_i}, {\text{W}^{K,\text{self},l}_i}, {\text{W}^{V,\text{self}, l}_i})_{i=1}^{H_{\text{Dec}}^{\text{self}}}$, each matrix being in $\mathbb{R}^{d_{H_\text{Dec}^\text{self}}\times d}$ where $d_{H_\text{Dec}}^{\text{self}} = d / H_\text{Dec}^{\text{self}}$
        \item A projection matrix $\text{W}^{O,\text{self}, l} \in \mathbb{R}^{d\times d}$
    \end{itemize}
    \item{\textbf{LayerNorm and Add:}}
    LayerNorm and the residual ``Add" operation are applied element-wise across each token's embedding dimension, i.e. column-wise when the input is structured as $d \times n$ where $d$ is the embedding dimension and $n$ is the sequence length.

    \item \textbf{Multi-Head Cross Attention:}
    \begin{itemize}
        \item $H^{\text{cross}}_{\text{Dec}}$ Cross Attention heads, parametrized by $({\text{W}^{Q, \text{cross}, l}_i}, {\text{W}^{K,\text{cross}, l}_i}, {\text{W}^{V, \text{cross},l}_i})_{i=1}^{H^{\text{cross}}_{\text{Dec}}}$, \\each matrix being in $\mathbb{R}^{d_{H_\text{Dec}^{\text{cross}}}\times d}$ where $d_{H^{\text{cross}}_\text{Dec}} = d / H^{\text{cross}}_\text{Dec}$
        \item A projection matrix $\text{W}^{O,\text{cross}, l} \in \mathbb{R}^{d\times d}$
    \end{itemize}
    \item{\textbf{LayerNorm and Add:}}
    LayerNorm and the residual ``Add" operation are applied element-wise across each token's embedding dimension, i.e. column-wise when the input is structured as $d \times n$ where $d$ is the embedding dimension and $n$ is the sequence length.
    
    \item{\textbf{Feedforward Network (FFN):}}
    $((\text{ReLU},\text{W}^{\text{FFN}, l}_{1}, \text{b}^{\text{FFN}, l}_{1}), (\text{Id},\text{W}^{\text{FFN}, l}_{2}, \text{b}^{\text{FFN}, l}_{2}))$.

    \item{\textbf{LayerNorm and Add:}}     LayerNorm and the residual ``Add" operation are applied element-wise across each token's embedding dimension, i.e. column-wise when the input is structured as $d \times n$ where $d$ is the embedding dimension and $n$ is the sequence length.
\end{enumerate}

We employ the T5-11B architecture from Raffel et al. \cite{t5} with $\textbf{Encoder}_{\text{fusion}}$ and $\textbf{Decoder}_{\text{Intention}}$.

\paragraph{$\textbf{Encoder}_{\text{fusion}}$}
$n_{\text{Enc}} = 24$ with a final dimension $d_{\text{T}}=1024$. In each layer $l$, the Multi-Head Self-Attention is made of $H_{\text{Enc}} = 128$ heads and the Feedforward Network has the following composition\\ $((\text{GeGLU},\text{W}^{\text{FFN}, l}_{1}, \text{b}^{\text{FFN}, l}_{1}), \allowbreak (\text{Id},\text{W}^{\text{FFN}, l}_{2}, \text{b}^{\text{FFN}, l}_{2}))$ with an inner dimension of 65536 i.e. $\text{W}^{\text{FFN}, l}_{1} \in \mathbb{R}^{65536\times 1024}$ and $\text{W}^{\text{FFN}, l}_{2} \in \mathbb{R}^{1024\times 65536}$. 

\paragraph{$\textbf{Decoder}_{\text{Intention}}$}
$n_{\text{Dec}} = 24$ with a final dimension $d_{\text{T}}=1024$. In each layer $l$, the Masked Multi-Head Self-Attention is made of $H^{\text{self}}_{\text{Dec}} = 128$ heads, the Multi-Head Cross-Attention is made of $H^{\text{cross}}_{\text{Dec}} = 128$ heads. The Feedforward Network $((\text{GeGLU},\text{W}^{\text{FFN},  l}_{1}, \allowbreak \text{b}^{\text{FFN}, l}_{1}), (\text{Id},\text{W}^{\text{FFN}, l}_{2}, \text{b}^{\text{FFN}, l}_{2}))$ has an inner dimension of 65536 i.e. $\text{W}^{\text{FFN}, l}_{1} \in \mathbb{R}^{65536\times 1024}$ and $\text{W}^{\text{FFN}, l}_{2} \in \mathbb{R}^{1024\times 65536}$.

\section{Mathematical Interpretations}
\subsection{Workflow Signal}

\subsubsection{Algebraic Foundations of Workflow Signals}

Let $\text{X}$ be a tensor space over the field of real numbers $\mathbb{R}$. 
We assume $\text{X}$ is a finite-dimensional real Hilbert space equipped with an inner product $\langle \cdot,\cdot \rangle: \text{X} \times \text{X} \to \mathbb{R}$, hence satisfying the following properties for all $x, y, z \in \text{X}$ and $\alpha \in \mathbb{R}$:

\begin{enumerate}
    \item Symmetry: $\langle x,y \rangle = \langle y,x \rangle$
    \item Linearity: $\langle \alpha x + y,z \rangle = \alpha\langle x,z \rangle + \langle y,z \rangle$
    \item Positive definiteness: $\langle x,x \rangle \geq 0$, with equality if and only if $x = 0$
\end{enumerate}

This inner product induces a norm:
\begin{equation*}
    \|.\|: \begin{cases}
        \text{X} \to \mathbb{R} \\
        x \mapsto \sqrt{\langle x,x \rangle}
    \end{cases}
\end{equation*}

Which in turn defines a metric:

\begin{equation*}
    d: \begin{cases}
        \text{X} \times \text{X} \to \mathbb{R} \\
        x, y \mapsto \|x-y\|
    \end{cases}
\end{equation*}

The completeness of $\text{X}$ with respect to this metric is guaranteed by our assumption that $\text{X}$ is a Hilbert space, meaning that every Cauchy sequence in $\text{X}$ converges to an element in $\text{X}$. Let $\{x_n\}_{n\in\mathbb{N}}$ a Cauchy sequence in $\text{X}$ ($\forall \varepsilon > 0, \exists N \in \mathbb{N}$ such that $\forall n, m > N, \|x_n - x_m\| < \varepsilon$), there exists $x \in \text{X}$ such that $\lim_{n\to\infty} \|x_n - x\| = 0$.\\

Given the finite dimensionality of $\text{X}$, we can construct an orthonormal basis $\{e_i\}_{i=1}^n$ of $\text{X}$, where $n = \dim(\text{X})$, satisfying:

\begin{enumerate}
    \item Orthonormality: $\forall i,j, \langle e_i,e_j \rangle = \delta_{ij}$ (where $\delta_{ij}$ is the Kronecker delta)
    \item Completeness: $\text{span}(\{e_i\}_{i=1}^n) = \text{X}$
\end{enumerate}

This orthonormal basis allows for the unique representation of any Workflow Signal thread vector $x \in \text{X}$ as a finite linear combination of $\{e_i\}_{i=1}^n$: 

\begin{equation*}
    \forall x \in \text{X}, \exists! (\alpha_i)_{i=1}^n \text{ such that } x = \sum_{i=1}^n \alpha_i e_i, \text{where } \alpha_i = \langle x,e_i \rangle 
\end{equation*}

The following aims to define a subspace $\text{S}$ of $\text{X}$ of Workflow Signals.\\

Let $\text{S}$ be a non-empty subspace of $\text{X}:\text{S} \subseteq \text{X}$. $\text{S}$ has the following properties:
\begin{enumerate}
    \item \text{Finite dimensionality: } $\dim(\text{S}) < \infty$
    \item \text{Inner product structure: } $\exists \langle \cdot, \cdot \rangle_{\text{S}}: \text{S} \times \text{S} \rightarrow \mathbb{R}$
    \item \text{Completeness: } $\text{S} \text{ is complete under } \|\cdot\|_{\text{S}} = \sqrt{\langle \cdot, \cdot \rangle_{\text{S}}}$
\end{enumerate}

Let $\text{I}$, $\text{P}$, $\text{O}$ be subspaces of $\text{S}$ corresponding to Input, Process and Output Workflow Signals:

\begin{equation}
\text{S} = \text{I} \oplus \text{P} \oplus \text{O}
\end{equation}

where $\oplus$ denotes the direct sum between two spaces, such that:
\begin{equation*}
\forall s \in \text{S}, \exists ! i \in \text{I}, p \in \text{P}, o \in \text{O},  \text{ such that } s = i + p + o
\end{equation*}
and
\begin{equation*}
\text{I} \cap \text{P} = \{0_\text{S}\}, \text{O} \cap \text{P} = \{0_\text{S}\} \text{ and } \text{I} \cap \text{O} = \{0_\text{S}\}
\end{equation*}

The assumption of the direct sum decomposition of \( \text{S} \) into the subspaces \( \text{I} \), \( \text{P} \), and \( \text{O} \) originates from the real world business context of Opus. The space \( \text{X} \) represents the context space, in which a set of Business Artefacts is encoded as one vector. The signal space \( \text{S} \subseteq \text{X} \) consists of Workflow Signals. In this framework, the Input representation of a Workflow Signal is determined by the context that defines it as an Input. While Input and Output may sometimes refer to the same underlying object, their representations remain distinct within a given context. For instance, a ``medical record'' can function as both an Input and an Output in a Process, but within a Workflow Signal describing a specific Workflow, the representation of ``medical record'' as an Input will differ from its representation as an Output. This distinction is context-dependent and ensures that Input and Output roles remain well-defined within the Workflow Signal space. Therefore we stipulate that each Workflow Signal can be uniquely decomposed into Input, Process and Output components. \\

We suppose that $\text{I}$, $\text{P}$, and $\text{O}$ are non empty.\\

As subspaces of $\text{S}$, we can define $\{e_{\text{I}, k}\}_{k=1}^{\text{dim($\text{I}$)}}$, $\{e_{\text{P}, k}\}_{k=1}^{\text{dim($\text{P}$)}}$, $\{e_{\text{O}, k}\}_{k=1}^{\text{dim($\text{O}$)}}$ orthonormal bases of $\text{I}$, $\text{P}$ and $\text{O}$ respectively.\\

We can define the projection operators $p_{\text{I}}: \text{S} \rightarrow \text{I}, p_{\text{P}}: \text{S} \rightarrow \text{P}$ \text{ and } $p_{\text{O}}: \text{S} \rightarrow \text{O}$ \text{such that:}
\begin{align*}
    &p_{\text{I}} + p_{\text{P}} + p_{\text{O}} = \text{Id}{\text{S}} \\
    &\forall s \in \text{S}, (p_{\text{I}} + p_{\text{P}} + p_{\text{O}})(s) = s \\
    &\text{Im}(p_{\text{I}}) + \text{Im}(p_{\text{P}}) + \text{Im}(p_{\text{O}}) = \text{S}
\end{align*}
\text{and}

\begin{equation*}
    p_{\text{I}}^2 = p_{\text{I}}, p_{\text{P}}^2 = p_{\text{P}}, p_{\text{O}}^2 = p_{\text{O}}
\end{equation*}

Based on the above, $s \in \text{S}$ can be decomposed as:
\begin{equation}
s = p_\text{I}(s) + p_\text{P}(s) + p_\text{O}(s)
\end{equation}

And each of $i$, $p$ and $o$ can be uniquely decomposed on their respective bases:

\begin{equation}
    s = \sum\limits_{k=1}^{\text{dim}(\text{I})} \alpha_{\text{I}, k} e_{\text{I}, k} + \sum\limits_{k=1}^{\text{dim}(\text{P})} \alpha_{\text{P}, k} e_{\text{P}, k} +
    \sum\limits_{k=1}^{\text{dim}(\text{O})} \alpha_{\text{O}, k} e_{\text{O}, k}
\end{equation}

This formalism enables the expression of generative families within the spaces I, P and O, which serve as the foundational idea for the class sets of the classification heads employed in the system.

\subsubsection{Generative families of \text{I}, \text{P}, \text{O}}
\label{sec:generativeFam}

\paragraph{Definition}
Let $\text{I}_g, \text{P}_g, \text{O}_g$ be generative families of $\text{I}, \text{P}, \text{O}$ respectively. \\
For $\text{X} \in \{\text{I}, \text{P}, \text{O}\}$,
$\text{X}_g = \{ e_{\text{X}_g, k}\}_k$, $\forall k, e_{\text{X}_g, k}\in \text{X}$
$\forall x \in \text{X}, \exists (\alpha_{\text{X}, k})_k$ such that 
$x = \sum\limits_{k=1}^{|\text{X}_g|} \alpha_{\text{X}, k} e_{\text{X}_g, k}$

The Input, Process and Output generative families can be built initially semantically using Large Language Models and iteratively updated from Workflow Signals as described in Algorithm 1.
\begin{algorithm}[H]
\caption{Adaptive Construction of Generative Family from Intention with Error Control}
\label{alg:adaptive_decomp}
\begin{algorithmic}[1]
\Require
    \State A Workflow Signal $x \in\text{X}, \text{X}\in\{\text{I}, \text{P}, \text{O} \}$
    \State Initial generative family:
        $\text{X}_g=\{e_{\text{X}_g,k}\}_{k=1}^{|\text{X}_g|}\subset \text{X},
        $
    \State Error threshold $\epsilon_{\text{X}} >0$
    \State Maximum iteration count ${\text{M}}_{{\text{max}}}$

\Ensure

Coefficient vectors $\boldsymbol{\alpha}_\text{X}\in\mathbb{R}^{|\text{X}_g|},\quad 
$ and sets such that
    \[ 
    \left\|x-\sum_{k=1}^{|\text{X}_g|}\alpha_{\text{X},k}\,e_{\text{X}_g,k}\right\| < \epsilon_\text{X}, 
    \]

\Function{DecomposeWithError}{$x, \text{X}_g, \epsilon, \text{M}_{max}$}
    \State $m \gets 0$
    \State $error \gets \infty$
    \While{$error > \epsilon$ AND $m < \text{M}_{max}$}
        \State \textbf{Solve} $\boldsymbol{\alpha}^*_\text{X} \gets \text{argmin}_{\boldsymbol{\alpha}_\text{X}} \left\|x-\sum_{k=1}^{|\text{X}_g|}\alpha_{\text{X}, k}\,e_{\text{X}_g,k}\right\|
        $
        \State $error \gets \left\|x-\sum_{k=1}^{|\text{X}_g|}\alpha^*_{\text{X}, k}\,e_{\text{X}_g,k}\right\|$
        \If{$error > \epsilon$}
            \State $\text{X}_g \gets \text{X}_g \cup \{x\}$
        \EndIf
        \State $m \gets m + 1$
    \EndWhile
    \Return $\boldsymbol{\alpha}^*_\text{X}, \text{X}_g$
\EndFunction

\Procedure{Main}{}
    \State $(\boldsymbol{\alpha}_\text{X}, \text{X}_g) \gets$ \Call{DecomposeWithError}{$x, \text{X}_g, \epsilon_\text{X}, \text{M}_\text{max}$}
\EndProcedure

\end{algorithmic}
\end{algorithm}

Algorithm 1 exhibits potentially large complexity: the worst-case iteration count of $\text{M}_{\text{max}}$ could be reached for each component, yielding $\mathcal{O}(3\text{M}_{\text{max}})$ iterations with each iteration solving an increasingly complex minimization problem. However, as the Input, Process and Output spaces are relatively constrained, a solution is reached within a bounded number of iterations and acceptable computational complexity.\\

Our system's parameter complexity and training cost are directly influenced by the size and stability of the generative families for Input, Process and Output classifications. Therefore, Algorithm 1 is systematically combined with a Gram–Schmidt-type algorithm to control the dimensionality of these families, which naturally expand when using Algorithm 1 alone. This process ensures convergence to a stable generative family structure. In practice, as we construct these generative families from granular signals \( i \), \( o \) and \( p \) (extracted from Business Artefacts), we enforce a rigorous dimensionality control mechanism—refining these families as new elements are incorporated.

\subsection{Workflow Intentions}

\subsubsection{Algebraic Foundations of Workflow Intention}
Let $\mathcal{G} = \text{I} \times \text{P} \times \text{O}$, $\gamma \in \mathcal{G}$ is an ordered triple representing a Workflow Intention in terms of Input, Output and Process Workflow Signals. 

\begin{equation*}
    \dim(\mathcal{G}) = \dim(\text{I}) + \dim(\text{P}) + \dim(\text{O})
\end{equation*}

The canonical projections on $\mathcal{G}$ are
\begin{equation}
    \begin{split}
    \pi_\text{I}&: \mathcal{G} \to \text{I}, \quad \forall \gamma = (i, p, o) \in \mathcal{G}: \pi_\text{I}(\gamma) = i\\
    \pi_\text{P}&: \mathcal{G} \to \text{P}, \quad \forall \gamma = (i, p, o) \in \mathcal{G}: \pi_\text{P}(\gamma) = p \\
    \pi_\text{O}&: \mathcal{G} \to \text{O}, \quad \forall \gamma = (i, p, o) \in \mathcal{G}: \pi_\text{O}(\gamma) = o
    \end{split}
\end{equation}

With kernels in $\mathcal{G}$:
\begin{equation}
    \begin{split}
\ker(\pi_\text{I}) &= \{\gamma \in \mathcal{G} : \gamma = (0_\text{I}, o, p)\}\\
\ker(\pi_\text{P}) &= \{\gamma \in \mathcal{G} : \gamma = (i, 0_\text{P}, o)\}\\
\ker(\pi_\text{O}) &= \{\gamma \in \mathcal{G} : \gamma = (i,p ,0_\text{O})\}
    \end{split}
\end{equation}

Finally, for a topology $\tau_\mathcal{G}$, $\forall U \in \tau_\mathcal{G}: U \subseteq \mathcal{G}$ and $\forall \gamma \in U$, there exist open neighborhoods $B_\text{I}(\mathcal{G}) \in \tau_\text{I}, B_\text{P}(\mathcal{G}) \in \tau_\text{P}$, and $B_\text{O}(\gamma) \in \tau_\text{O}$ such that $\gamma \in (B_\text{I}(\gamma) \times B_\text{P}(\gamma) \times B_\text{O}(\gamma)) \subseteq U$. This allows the Opus system to understand ``closeness" of Workflow Intentions, that is, if something is ``close" in $\mathcal{G}$, it is close in all three components ($\text{I}$, $\text{P}$, and $\text{O}$) simultaneously.\\

A single Workflow Intention $\gamma$ in $\mathcal{G}$ is defined as:

\begin{equation}
    \gamma = (i, p, o) \in \mathcal{G} \text{ such that } i \in \text{I}, p \in \text{P} \text{ and } o \in \text{O}
\end{equation}

Workflow Intentions can be combined to accommodate users with hybrid Workflow Intention:
\begin{equation}
        \gamma_1 + \gamma_2 = (i_1,p_1,o_1) + (i_2,p_2,o_2) = (i_1+i_2, p_1+p_2, o_1+o_2)
\end{equation}

Workflow Intention representations can have different strengths i.e. expressiveness in different contexts

\begin{equation}
\alpha\gamma = \alpha(i,p,o) = (\alpha i, \alpha p, \alpha o)
\end{equation}

\subsubsection{Workflow Intentions from Workflow Signal}

Let $f: \text{X}\times \text{S} \mapsto \mathcal{P}(\mathcal{G})$ the powerset of $\mathcal{G}$. By definition, 
$\mathcal{P}(\mathcal{G}) = \{A \mid \forall x \in A, x \in \mathcal{G}\}$.\\ 

Let $(x, s) \in \text{X} \times \text{S}$,
\begin{align}
 &\exists n\in \mathbb{N}, \Gamma = \{\gamma_k\}_{k=1}^n \text{ such that } f(x, s) = \Gamma \text{ and } \\ 
 &\forall \gamma \in \Gamma, \gamma \in \mathcal{G}
\text{ and } \exists i_{\gamma} \in \text{I}, p_{\gamma} \in \text{P}, o_{\gamma} \in \text{O} \text{ such that } \gamma = (i_{\gamma}, p_{\gamma}, o_{\gamma}) \notag
\end{align}

\paragraph{Property: Information Conservation}
Let $x \in \text{X}, s \in \text{S}$ and $\Gamma = f(x,s)$.
\begin{equation}
    \forall \gamma \text{ in } f(x, s), \quad \|i_{\gamma}\| \leq \|i_{s}\|, \|p_{\gamma}\| \leq \|p_{s}\| \text{ and } \|o_{\gamma}\| \leq \|o_{s}\|
\end{equation}

Any Workflow Signal (Input, Process or Output) of a Workflow Intention object of a Workflow Intention Set is weaker or equal than the overall Workflow Signal (Input, Process or Output) the Workflow Intention Set was derived from.

\paragraph{Property: Workflow Intention Variation}
Let $x \in \text{X}, s \in \text{S}$ and $\Gamma = f(x,s)$, if $|f(x, s)| > 1$, 
\begin{align}
    &\forall \gamma_1, \gamma_2 \in f(x, s), \Bigg|\Big\{v ,  \frac{\langle v_{\gamma_1}, v_{\gamma_2} \rangle}{\|v_{\gamma_1}\| \|v_{\gamma_2}\|} < \epsilon_{\text{sim}}, v \in \{i, p, o\}\Big\}\Bigg| \geq 1, &\epsilon_{\text{sim}} \in [0,1[ \\
    &\forall \gamma_1, \gamma_2 \in f(x, s), \Bigg|\Big\{v ,\|v_{\gamma_1}-v_{\gamma_2}\| > \epsilon_2, v \in \{i, p, o\}\Big\}\Bigg| \geq 1, &\epsilon_2 \in [0,1[ 
\end{align}

If the Workflow Intention Set is composed of two or more Workflow Intention objects, each pair of Workflow Intention must present variation on at least one Workflow Signal dimension (Input, Process or Output). 

\end{document}